\documentclass[letterpaper,11pt]{article}

\usepackage[utf8]{inputenc}
\usepackage[T1]{fontenc}
\usepackage{times}
\usepackage[margin=1in,top=0.45in]{geometry}
\usepackage{microtype}

\usepackage{amsmath,amssymb,amsfonts,amsthm}
\usepackage{mathrsfs}
\usepackage{dsfont}
\usepackage{nicefrac}

\usepackage{graphicx}
\usepackage{tabularx}
\usepackage{longtable}

\usepackage{blindtext}
\usepackage{etoc}

\PassOptionsToPackage{numbers, sort&compress}{natbib}
\usepackage{natbib}
\usepackage{url}
\usepackage{fontawesome5} 

\usepackage{amsmath,amsfonts,bm}

\def\eqref#1{equation~\ref{#1}}

\def\1{\bm{1}}

\DeclareMathAlphabet{\mathsfit}{\encodingdefault}{\sfdefault}{m}{sl}
\SetMathAlphabet{\mathsfit}{bold}{\encodingdefault}{\sfdefault}{bx}{n}

\newcommand{\method}{{\fontfamily{lmtt}\selectfont \textbf{TextReg}}\xspace}

\usepackage[textsize=tiny]{todonotes}
\usepackage[algo2e]{algorithm2e}
\usepackage[export]{adjustbox}
\usepackage{threeparttable}
\usepackage{multirow}
\usepackage{enumitem}
\usepackage{newfloat}
\usepackage{listings}
\usepackage{colortbl}
\usepackage{amsfonts}
\usepackage{amssymb}
\usepackage{pifont}
\usepackage{booktabs}
\usepackage{float}
\usepackage{bm}
\usepackage{tabulary}
\usepackage{makecell}
\usepackage{bbm}
\usepackage{mdframed}
\usepackage{algorithm}
\usepackage{algorithmic}
\usepackage{wrapfig}
\usepackage{caption,subcaption}
\makeatletter
\@ifpackageloaded{hyperref}{}{\usepackage[pagebackref=false,breaklinks=true,colorlinks,bookmarks=false]{hyperref}}
\makeatother
\usepackage[most]{tcolorbox}
\tcbuselibrary{theorems}
\definecolor{lightgray}{gray}{.9}
\definecolor{deepgray}{gray}{.8}
\tcbset{highlight math/.append style={left=0mm,right=0mm,top=0mm,bottom=0mm, colframe=white}}

\usepackage[capitalize,noabbrev]{cleveref}

\newcolumntype{I}{!{\vrule width 1pt}}
\makeatletter
\newcommand{\thickhline}{%
    \noalign {\ifnum 0=`}\fi \hrule height 1pt
    \futurelet \reserved@a \@xhline
}
\makeatother

\crefname{proposition}{Prop.}{Props.}
\crefname{section}{Sec.}{Secs.}
\crefname{table}{Tab.}{Tabs.}

\usepackage{xspace}
\makeatletter
\DeclareRobustCommand\onedot{\futurelet\@let@token\@onedot}
\def\@onedot{\ifx\@let@token.\else.\null\fi\xspace}

\usepackage[dvipsnames]{xcolor}
\usepackage{colortbl}
\definecolor{ada_blue}{rgb}{0,205,205}
\definecolor{glt_red}{rgb}{109,205,255}
\definecolor{MorandiBlue}{RGB}{118,134,146}
\definecolor{demphcolor}{RGB}{144,144,144}
\definecolor{mygray}{gray}{0.4}
\definecolor{autopurple}{HTML}{7030A0}
\definecolor{dyna_yellow}{HTML}{BF9000}
\definecolor{adaptive_blue}{HTML}{0070C0}
\definecolor{darkgrey}{RGB}{120,120,120}
\definecolor{mygrey}{RGB}{200,200,200}
\definecolor{myblue}{HTML}{00CDCD}
\definecolor{champagne}{rgb}{0.97, 0.91, 0.81}
\definecolor{darksalmon}{rgb}{0.91, 0.59, 0.48}
\definecolor{emerald}{rgb}{0.31, 0.78, 0.47}
\definecolor{green(pigment)}{rgb}{0.0, 0.65, 0.31}
\definecolor{amaranth}{rgb}{0.9, 0.17, 0.31}
\definecolor{iris}{rgb}{0.35, 0.31, 0.81}
\definecolor{uu}{rgb}{0.95, 0.51, 0.51}
\definecolor{spirodiscoball}{rgb}{0.06, 0.75, 0.99}
\definecolor{cadetblue}{RGB}{95,158,160}
\definecolor{keywordcolor}{RGB}{178,34,34}

\definecolor{customgreen}{HTML}{667b5b}
\definecolor{customblue}{HTML}{bcccea}

\usepackage{pifont}
\usepackage{bbding}

\SetCommentSty{mycommfont}

\newcommand{\reddown}[1]{_{\color{darksalmon}\downarrow #1}}
\newcommand{\greenup}[1]{_{\color{green(pigment)}\uparrow #1}}

\crefname{equation}{Eq.}{Eqs.}
\Crefname{equation}{Equation}{Equations}

\theoremstyle{definition}

\renewcommand{\thefootnote}{\fnsymbol{footnote}}

\setlist{leftmargin=5mm}

\definecolor{abstractbg}{RGB}{230,242,250}
\definecolor{abstractborder}{RGB}{200,210,220}
\usepackage{titlesec}

\titleformat{\section}
  {\sffamily\bfseries\Large}
  {\thesection}
  {1em}
  {}

\renewenvironment{abstract}
{
\begin{center}
\begin{tcolorbox}[
    colback=abstractbg,
    colframe=abstractborder,
    boxrule=0.6pt,
    arc=6pt,
    width=\textwidth,
    left=8pt,
    right=8pt,
    top=8pt,
    bottom=8pt
]
}
{
\end{tcolorbox}
\end{center}
}

\newcommand{\papertitle}{%
\sffamily\bfseries\fontsize{16}{1}\selectfont
TextReg: Mitigating Prompt Distributional Overfitting \\[.35em] via Regularized Text-Space Optimization
}

\newcommand{\paperauthors}{%
\sffamily
Lucheng Fu$^{1}$, Ye Yu$^{2}$, Yiyang Wang$^{1}$, Yiqiao Jin$^{1}$,
\\[0.35em]
Haibo Jin$^{2}$, B.~Aditya Prakash$^{1\dagger}$, Haohan Wang$^{2\dagger}$
\\[0.55em]
{\normalsize $^{1}$Georgia Institute of Technology \qquad $^{2}$University of Illinois Urbana-Champaign}
\\[0.5em]
{\normalsize \faGlobe~Website: \url{https://textreg.github.io/}}
\\[0.15em]
{\normalsize \faGithub~GitHub: \url{https://github.com/luchengfu6/TextReg}}
\footnotetext[2]{Corresponding authors.\newline\hspace*{1.8em}Contact: \texttt{luchengfu@gatech.edu}}
}

\newcommand{\paperdate}{}

\begin{document}

\thispagestyle{empty}

\noindent
\includegraphics[height=.8cm]{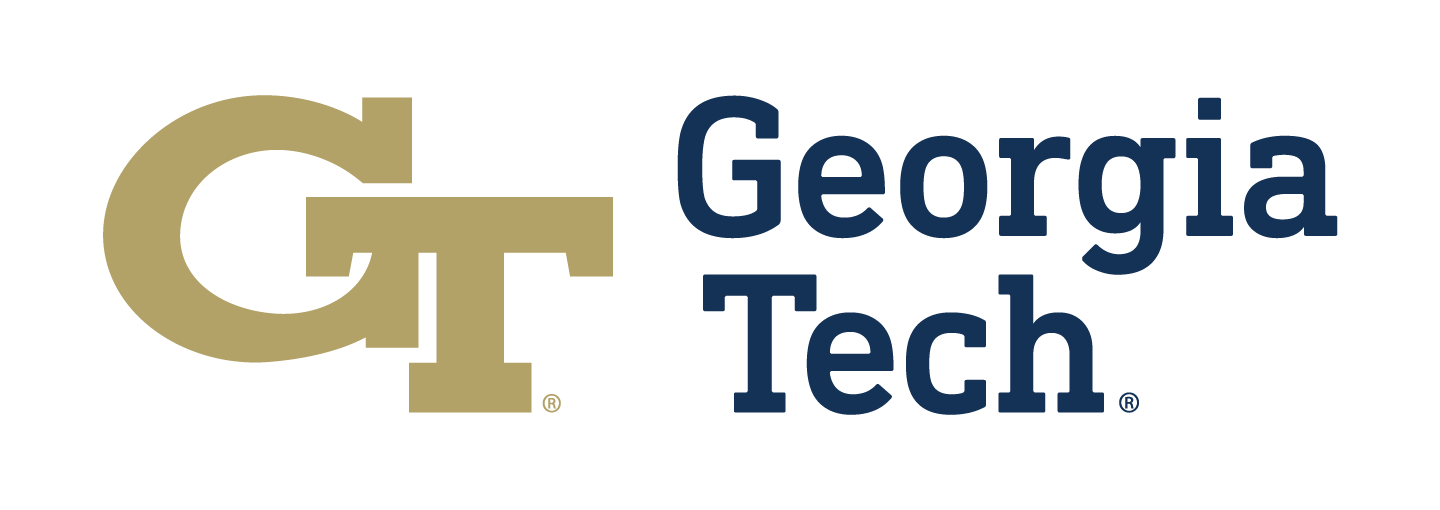}
\includegraphics[height=.8cm]{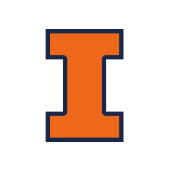}

\vspace{-.15in}

\noindent\rule{\textwidth}{0.8pt}

\vspace{0.65cm}

\begin{center}
    {\papertitle\par}
    \vspace{0.45cm}
    {\large \paperauthors\par}
    \vspace{0.2cm}
    {\normalsize \paperdate\par}
\end{center}

\begin{abstract}
Large language models (LLMs) are highly sensitive to the prompts used to specify task objectives and behavioral constraints. Many recent prompt optimization methods iteratively rewrite prompts using LLM-generated feedback, but the resulting prompts often become longer, accumulate narrow sample-specific rules, and generalize poorly beyond the training distribution. We study this failure mode as \emph{prompt distributional overfitting} and argue that it reflects a lack of representation control in discrete text-space optimization. We formalize this view through \emph{representational inefficiency}, 
a dual-factor measure that decomposes prompt inefficiency into 
capacity cost and scope narrowness, attributing distributional prompt 
overfitting to their coupled growth during optimization. We propose \method{}, a regularization framework that realizes a soft-penalty objective through regularized textual gradients, combining Dual-Evidence Gradient Purification, Semantic Edit Regularization, and Regularization-Guided Prompt Update. Across multiple reasoning benchmarks, \method{} substantially improves out-of-distribution (OOD) generalization, with accuracy gains of up to +11.8\% over TextGrad and +16.5\% over REVOLVE.
\end{abstract}

\setcounter{footnote}{0}
\renewcommand{\thefootnote}{\arabic{footnote}}

\section{Introduction}
\label{sec:intro}
Large language models (LLMs) have exhibited strong performance on a wide variety of reasoning and generation tasks~\citep{brown2020language, achiam2023gpt, team2023gemini, grattafiori2024llama}, with the input prompt serving as the central interface that specifies task objectives, output formats, and behavioral constraints~\citep{wang2026mascot,wang2025companioncast}. This sensitivity to prompting has motivated the development of \emph{prompt optimization}~\citep{pryzant2023automatic,wan2024teach}, which aims to automatically refine prompts through data-driven structure discovery and feedback, thereby reliably eliciting desired behavior and reducing manual trial-and-error in prompt design. Recent approaches iteratively refine prompts using natural-language feedback from LLM evaluators~\citep{pryzant2023automatic,yuksekgonul2024textgrad} and have achieved promising empirical improvements. Despite these gains, optimized prompts often suffer from poor generalization beyond the training data: as optimization proceeds, prompts may become longer, accumulate case-specific instructions, and become more sensitive to input variations~\citep{zhuo2024prosa}, a failure mode known as \emph{prompt distributional overfitting}~\citep{khattak2023self}. This failure manifests along two coupled dimensions: prompts not only expand in length as new rules, exceptions, examples, and stylistic constraints are appended, but also narrow in scope as the appended content drifts toward fragmented sample-dependent patches rather than compact general principles. Existing prompt optimizers provide little control over either dimension.

We view a prompt as a structured representation of task knowledge, composed of behavioral rules expressed in natural language; an ideal prompt should encode broadly applicable principles in a compact form, rather than fit training examples through an expanding set of special-case instructions. We formulate this failure mode through the lens of \emph{representational inefficiency}: from this perspective, unconstrained optimization violates this principle in two coupled ways. First, increasing prompt length raises the \emph{capacity cost}---longer prompts consume more context budget and make it harder for the model to reliably locate and leverage relevant instructions, given that LLMs' effective context window is limited~\cite{levy2024same,liu2024lost,jin2025sara}. Second, prompts can suffer from \emph{rule narrowing}: they accumulate rules that reduce training loss but apply only to a narrow subset of inputs, functioning as ad hoc patches rather than reusable task principles. These two factors jointly lead to representational inefficiency, arising from the interaction between the capacity cost incurred by prompt length and the scope waste among its constituent rules. As they compound, an increasing fraction of the prompt's capacity is wasted on information that contributes little to generalization, suggesting that prompt distributional overfitting~\citep{khattak2023self} is fundamentally a failure of representation efficiency, rather than merely an artifact of optimization dynamics.

\definecolor{P1}{RGB}{191, 144, 0}    
\definecolor{P2}{RGB}{14, 132, 140}   

\begin{wrapfigure}{r}{0.5\textwidth}
    \vspace{-20pt}
    \begin{center}
        \includegraphics[width=0.5\textwidth]{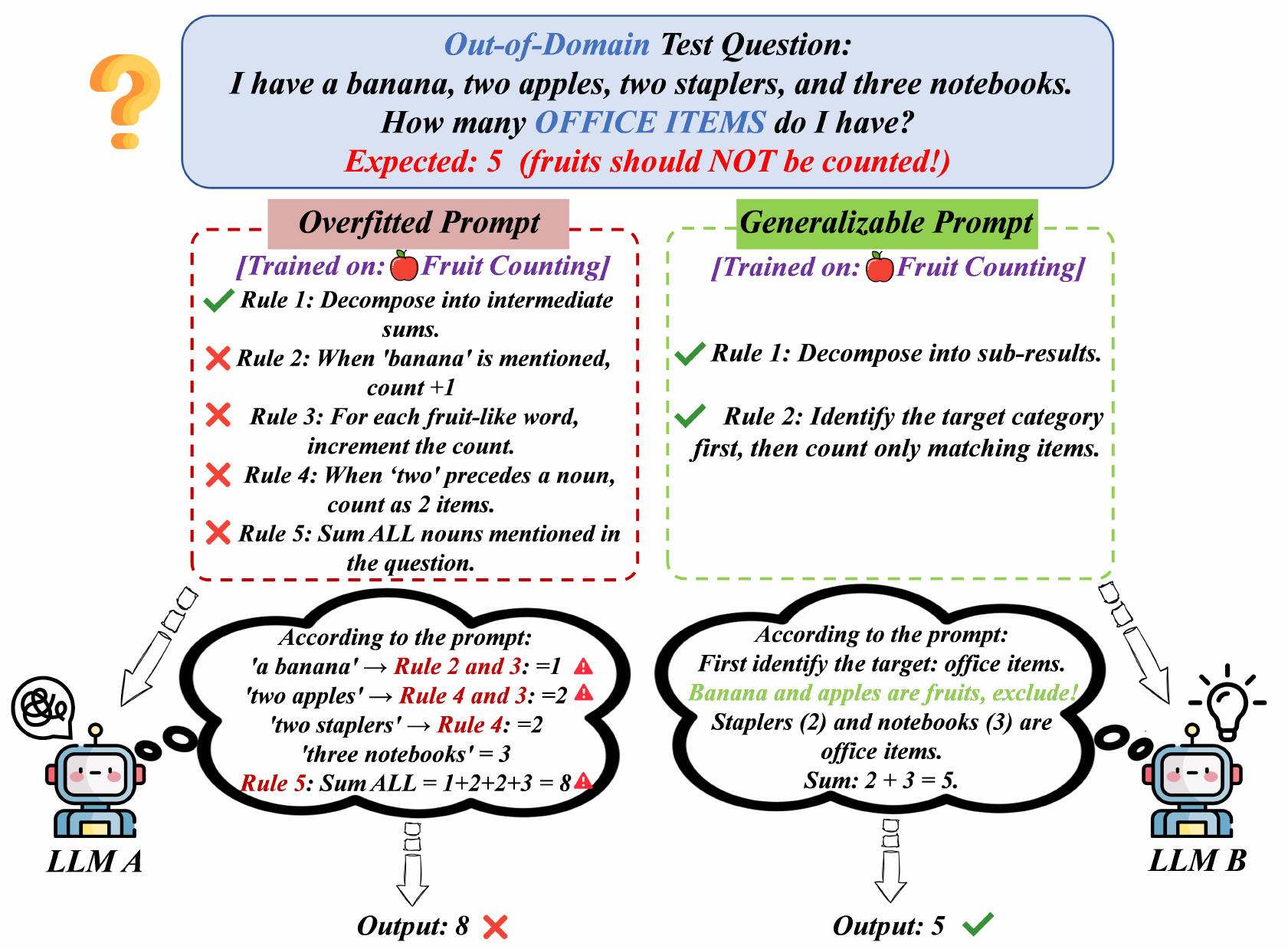}
    \end{center}
    \vspace{-10pt}
    \caption{\small \textbf{Problem Illustration.} We illustrate prompt distributional overfitting in prompt optimization:
    \textbf{\uppercase\expandafter{\romannumeral1})} conventional methods often produce \textcolor{P1}{long prompts saturated with narrow rules} (left), which \textbf{degrade on OOD inputs}.
    \textbf{\uppercase\expandafter{\romannumeral2})} Our goal is to instead yield \textcolor{P2}{compact prompts composed of broadly applicable rules} (right), achieving \textbf{stronger OOD generalization}.}
    \label{fig:problem}
    \vspace{-15pt}
\end{wrapfigure}

In classical machine learning, overfitting is commonly mitigated by regularization that controls model complexity.
However, applying this perspective to prompt optimization is non-trivial: the optimization is non-differentiable, with feedback generated by LLM evaluators and updates produced by LLM rewriting.

To address this challenge, we propose \method, a regularization framework that improves prompt optimization through a regularized textual-gradient view: prompt updates should follow both a task direction that improves empirical performance and a regularization direction that constrains the growth of representational inefficiency. \method{} realizes this view through three stages. \emph{Dual-Evidence Gradient Purification} constructs a purified task gradient by combining local batch evidence and RuleBank recurrence evidence to distinguish case-specific patches from broadly applicable rules. \emph{Semantic Edit Regularization} estimates how recent prompt edits change the capacity cost and scope waste, converting the observed degradation into a textual regularization gradient. \emph{Regularization-Guided Prompt Update} then uses this gradient to steer prompt rewriting toward task-consistent edits that avoid unnecessary increases in representational inefficiency. Our contributions are threefold:






\begin{itemize}[leftmargin=*]

\item[\ding{182}] \textbf{Formalizing Representational Inefficiency.}
We formulate prompt distributional overfitting as a failure of representation efficiency in discrete text-space optimization, and introduce \emph{representational inefficiency}---a dual-factor measure capturing the interaction between capacity cost from prompt length growth and scope waste from rule narrowing.

\item[\ding{183}] \textbf{Regularized Text-Space Prompt Optimization.}
We propose \method{}, a regularization framework that realizes a soft-penalty objective in discrete text space through three complementary stages---Dual-Evidence Gradient Purification, Semantic Edit Regularization, and Regularization-Guided Prompt Update. \method{} explicitly controls the growth of representational inefficiency during optimization, producing prompts that generalize beyond the training distribution.

\item[\ding{184}] \textbf{Improved Out-of-Distribution Generalization.}
Through extensive experiments across multiple reasoning benchmarks, we demonstrate that \method{} consistently outperforms existing prompt optimization methods on out-of-distribution (OOD) generalization across both datasets and test engines, validating that controlling representational inefficiency mitigates prompt distributional overfitting.

\end{itemize}

\section{Related Work}
\label{sec:related}
\paragraph{Prompt optimization.}
Reasoning-oriented prompting techniques such as CoT~\citep{wei2022chain}, self-consistency~\citep{wang2022self}, ReAct~\citep{yao2022react}, PoT~\citep{chen2022program}, and ToT~\citep{yao2023tree} improve LLM reasoning by structuring the inference process. Automated prompt optimization discovers effective prompts algorithmically. Approaches range from discrete token- or edit-based methods such as AutoPrompt~\citep{shin2020autoprompt} and RLPrompt~\citep{deng2022rlprompt} to LLM-based methods such as APE~\citep{zhou2022large}, EvoPrompt~\citep{guo2023connecting}, Promptbreeder~\citep{fernando2023promptbreeder}, and DSPy~\citep{khattab2023dspy}.
The most relevant methods use feedback from LLM evaluators to guide iterative rewriting.
APO~\citep{pryzant2023automatic} treats LLM-generated critiques as discrete-space gradients, TextGrad~\citep{yuksekgonul2024textgrad} formalizes textual differentiation through arbitrary computational graphs, and REVOLVE~\citep{zhang2024revolve} tracks response evolution across optimization steps. SIPDO~\citep{yu2025sipdo} complements this direction by coupling prompt optimization with synthetic data generation.
Although these methods have shown empirical gains, they primarily use task feedback to improve performance and do not explicitly regularize representational inefficiency during optimization.

\paragraph{Prompt robustness and overfitting.}
Recent work shows that prompt-based systems can be fragile under wording changes, semantic perturbations, adversarial paraphrases, and distribution shifts, as well as in long-context settings~\citep{zhuo2024prosa, levy2024same, liu2024lost}. Robust prompt optimization methods address related issues by accounting for distribution shifts~\citep{li2023robust} or using sharpness-aware prompt evolution to improve paraphrase invariance~\citep{wan2025beyond}.
However, prompt overfitting remains a distinct failure mode in feedback-based optimization: prompts may improve on training feedback while accumulating narrow, sample-specific instructions that fail to transfer. 
APO~\citep{pryzant2023automatic} and TextGrad~\citep{yuksekgonul2024textgrad} report this mismatch between training improvements and generalization. DLPO~\citep{peng2025dlpo} mitigates overfitting through static simplification instructions, while REMO~\citep{wu2025reflection} uses external memory and meta-reflection. \method{} treats prompt overfitting as a problem of controlling prompt representations: we formalize it through representational inefficiency and regularize optimization trajectories based on observed structural changes.

\paragraph{Regularization for generalization.}
Regularization can improve generalization in machine learning by controlling effective model capacity. Examples include weight decay~\citep{hoerl1970ridge, krogh1991simple}, dropout~\citep{wan2013regularization, srivastava2014dropout}, LASSO and elastic net~\citep{tibshirani1996regression, zou2005regularization}, and early stopping~\citep{caruana2000overfitting}.
Continuous prompt learning offers another way to limit the parameters optimized for a task: soft-prompt and prompt-tuning methods learn prompt embeddings and have been studied in few-shot settings~\citep{li2021prefix, lester2021power, liu2022p}.
\method{} regularizes natural-language prompts by jointly controlling capacity cost and scope narrowness, rather than constraining model parameters, continuous prompt embeddings, or prompt length alone.

\section{Problem Formulation}
\label{sec:problem}

\subsection{Problem Setup and Notation}

We consider a black-box language model $\mathcal{M}$ that maps an input $x$ and a textual prompt $p$ to an output $\mathcal{M}(p,x)$. Let $\mathcal{P}$ denote the space of valid prompts. Given a dataset $\mathcal{D}=\{(x_i,y_i)\}_{i=1}^N$, we define the empirical task risk of a prompt as
\begin{equation}
    \mathcal{L}_{\mathcal{D}}(p)
    =
    \frac{1}{|\mathcal{D}|}
    \sum_{(x,y)\in\mathcal{D}}
    \ell\big(\mathcal{M}(p,x),y\big),
    \label{eq:empirical-risk}
\end{equation}
where $\ell(\cdot,\cdot)$ is a task-specific loss, possibly induced by an evaluator in generative settings. Standard prompt optimization seeks a prompt $p\in\mathcal{P}$ that minimizes $\mathcal{L}_{\mathcal{D}_{\text{train}}}(p)$.

\subsection{Prompt Distributional Overfitting and Generalization}

Optimizing~\cref{eq:empirical-risk} does not guarantee generalization beyond the training distribution. For a target distribution $\mathcal{D}'$ that differs from $\mathcal{D}_{\text{train}}$, we define the generalization gap as
\begin{equation}
    \Delta(p;\mathcal{D}')
    =
    \mathcal{L}_{\mathcal{D}'}(p)
    -
    \mathcal{L}_{\mathcal{D}_{\text{train}}}(p).
\end{equation}

\paragraph{Prompt Distributional Overfitting.}
Prompt distributional overfitting occurs when optimization reduces
$\mathcal{L}_{\mathcal{D}_{\text{train}}}(p)$ while increasing
$\Delta(p;\mathcal{D}')$. Empirically, this behavior is often accompanied by structural growth of the prompt, including increased length and the accumulation of narrow, sample-specific instructions.

\paragraph{Connection to classical overfitting.}
This phenomenon parallels classical overfitting in that additional representational capacity can fit increasingly narrow patterns at the expense of generalization. We focus specifically on OOD generalization: an optimized prompt may continue to perform well on held-out samples from the source distribution while degrading on harder task variants or related-but-shifted tasks.

\subsection{Representational Inefficiency}

For analysis, we view a prompt $p$ as encoding a set of semantically distinct behavioral directives
$R(p)=\{r_1,\ldots,r_k\}$. A directive may describe a reasoning procedure, behavioral constraint, or other task-relevant instruction; this decomposition is analytical and does not require the prompt itself to be written as an explicit list of rules.

\paragraph{Generalization scope.}
Although optimization is performed on $\mathcal{D}_{\text{train}}$, we want the resulting prompt to remain useful over an input space $\mathcal{X}$ that preserves the underlying task while varying in surface form, scale, or difficulty. For a directive $r_i$, we define its generalization scope as
\begin{equation}
    s(r_i)
    \triangleq
    \mathbb{E}_{x\sim\mathcal{X}}
    \big[\rho_{r_i}(x)\big]
    \in[0,1],
\end{equation}
where $\rho_{r_i}(x)$ measures the degree to which $r_i$ is relevant to input $x$. Let
$\bar{s}(p)=\frac{1}{|R(p)|}\sum_{r_i\in R(p)}s(r_i)$ denote the average scope of the prompt.

For example, in object counting, a general procedure that applies across inputs has high scope, whereas an instruction tied to a particular entity has low scope:
\vspace{-1mm}
\begin{tcolorbox}[
    enhanced,
    colframe=black!70,
    colback=yellow!5,
    boxrule=1pt, arc=4mm,
    left=2mm, right=2mm, top=1mm, bottom=1mm,
]
\textbf{\textcolor{keywordcolor}{\ding{182} Broad rule}}:
\textit{List each item in the input one by one before producing the final count.} \\[1ex]
\textbf{\textcolor{keywordcolor}{\ding{183} Narrow rule}}:
\textit{When the input mentions tomatoes, count them as vegetables, not fruits.}
\end{tcolorbox}

\paragraph{Inefficiency measure.}
We define the representational inefficiency of a prompt as
\begin{equation}
    \mathcal{I}(p)
    =
    \underbrace{|p|_{\text{tok}}}_{\text{capacity cost }C(p)}
    \cdot
    \underbrace{\big(1-\bar{s}(p)\big)}_{\text{scope narrowness }W(p)}.
    \label{eq:representation-inefficiency}
\end{equation}
Here, $C(p)$ captures the context capacity consumed by the prompt, while
$W(p)$ captures the degree to which its directives fail to provide broadly applicable guidance. The multiplicative form emphasizes their interaction: additional prompt capacity is increasingly inefficient when it is devoted to narrowly applicable instructions. Thus, $\mathcal{I}(p)$ is high when a prompt is simultaneously long and dominated by low-scope directives.

\subsection{Regularized Text-Space Prompt Optimization}

Standard prompt optimization minimizes~\cref{eq:empirical-risk} without explicitly controlling how the prompt representation evolves. To formalize the desired trade-off, we consider the idealized regularized objective
\begin{equation}
    \min_{p\in\mathcal{P}}
    \quad
    \mathcal{L}_{\mathcal{D}_{\text{train}}}(p)
    +
    \lambda\,\mathcal{I}(p),
    \label{eq:regularized-objective}
\end{equation}
where $\lambda\geq0$ conceptually controls the trade-off between task performance and representational efficiency.

Importantly, \cref{eq:regularized-objective} should be understood as a conceptual regularization objective that characterizes the desired trade-off between task performance and representational efficiency. In discrete text space, however, its two structural components have different operational properties: capacity cost is directly observable through prompt length, whereas scope narrowness is semantic and depends on how broadly individual directives apply across the task distribution. This distinction motivates treating representational inefficiency through complementary structural signals rather than as a single directly measurable scalar.

This formulation therefore provides a principled view of what should be controlled during prompt optimization, while leaving the concrete realization of these controls to the optimization procedure introduced next.

\section{Method}
\label{sec:method}

\subsection{Overview}
\label{subsec:method-overview}

Rather than numerically optimizing~\cref{eq:regularized-objective}, \method{} operationalizes its task and regularization components through natural-language update signals. At optimization step $t$, it constructs a purified task signal $\tilde{g}_{\text{task}}$ and, when structural degradation is detected, a regularization signal $g_{\text{reg}}$. The next prompt is produced by an LLM-based rewrite operator
\begin{equation}
    p_{t+1}
    =
    \mathcal{U}
    \big(
        p_t;\,
        \tilde{g}_{\text{task}},
        g_{\text{reg}}
    \big),
    \label{eq:textreg-update}
\end{equation}
where $\mathcal{U}$ integrates the two signals in natural language rather than through numerical gradient addition. We use ``gradient'' in the textual-optimization sense: these signals specify update directions in language and are not numerical derivatives or vectors.

\begin{figure*}[t]
    \centering
    \includegraphics[width=0.96\textwidth]{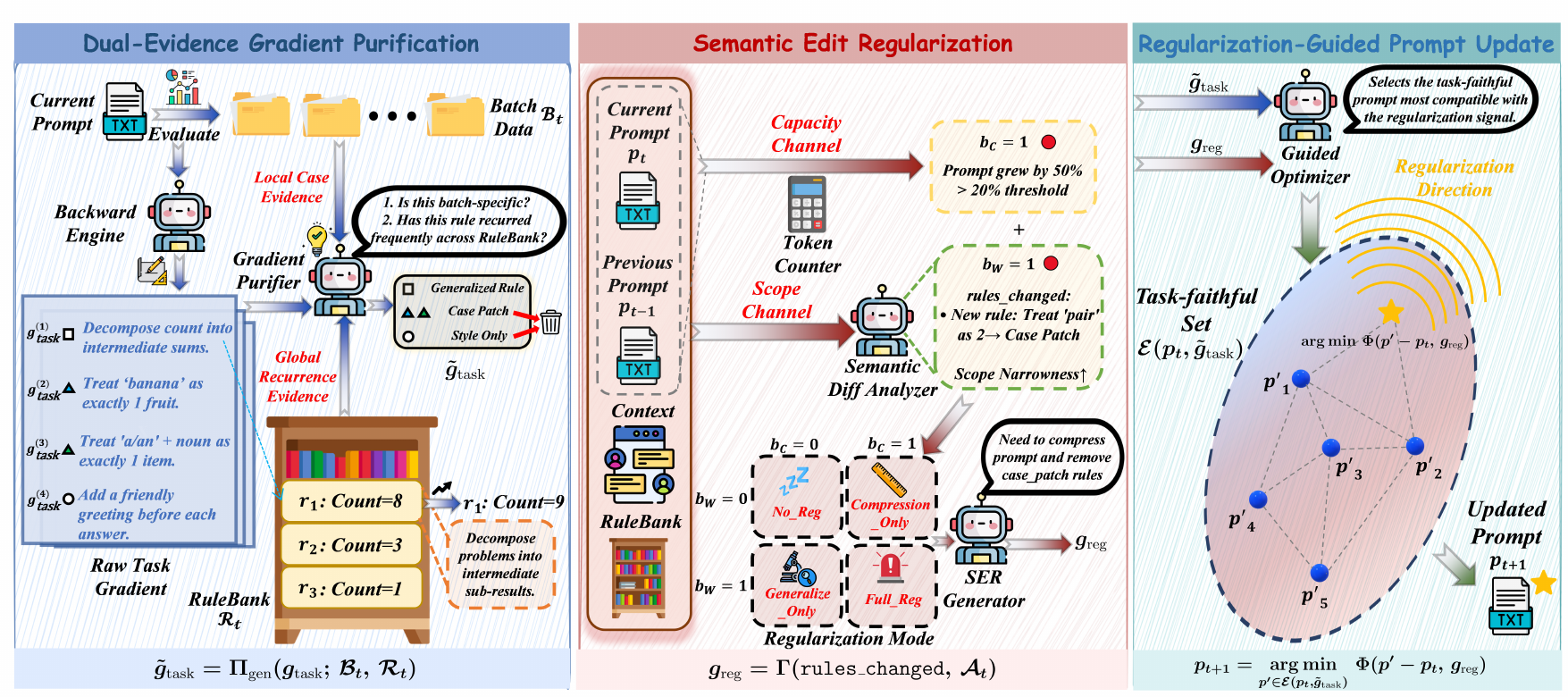}
    \caption{
    Overview of \method{}, which proceeds in three stages.
    (a) \textbf{\textit{Left}}: Dual-Evidence Gradient Purification filters the raw task gradient using local batch and RuleBank recurrence evidence, yielding $\tilde{g}_{\text{task}}$.
    (b) \textbf{\textit{Middle}}: Semantic Edit Regularization detects capacity and scope degradation and synthesizes the regularization signal $g_{\text{reg}}$.
    (c) \textbf{\textit{Right}}: Regularization-Guided Prompt Update rewrites $p_t$ into $p_{t+1}$ using both task and regularization guidance.
    }
    \label{fig:framework}
    \vspace{-6mm}
\end{figure*}

As illustrated in~\cref{fig:framework}, \method{} proceeds through three complementary stages.
\textbf{Stage~1 (Dual-Evidence Gradient Purification)} filters raw task gradients using local batch evidence and global RuleBank recurrence evidence.
\textbf{Stage~2 (Semantic Edit Regularization)} monitors the capacity and scope channels of representational inefficiency and constructs $g_{\text{reg}}$ when degradation is detected.
\textbf{Stage~3 (Regularization-Guided Prompt Update)} coordinates task and regularization guidance while preserving task-faithful corrections.

The key design is to control representational inefficiency at both the
\emph{feedback} and \emph{edit} levels. Stage~1 acts before rewriting by
suppressing feedback likely to produce narrow or redundant instructions,
whereas Stage~2 inspects the realized prompt transition and corrects
capacity or scope degradation that still emerges during rewriting.
Stage~3 then mediates these structural corrections without sacrificing
task-relevant updates. This separation allows \method{} to regulate both
how optimization signals are formed and how they are ultimately expressed
in the prompt, directly targeting the two sources of inefficiency,
$C(p)$ and $W(p)$.

\subsection{Dual-Evidence Gradient Purification}
\label{subsec:gradient-purification}

Prompt optimization can introduce narrow corrections tied to particular training examples, increasing scope narrowness $W(p)$, as well as redundant or stylistic updates that increase capacity cost $C(p)$ without adding useful behavior. Dual-Evidence Gradient Purification filters such feedback before it enters the prompt rewrite.

For a task gradient $g_{\text{task}}$ produced from mini-batch
$\mathcal{B}_t\subset\mathcal{D}_{\text{train}}$, we write
\begin{equation}
    \tilde{g}_{\text{task}}
    =
    \Pi_{\text{gen}}
    \big(
        g_{\text{task}};\,
        \mathcal{B}_t,\,
        \mathcal{R}_t
    \big),
    \label{eq:dual-evidence-projection}
\end{equation}
where $\Pi_{\text{gen}}$ is an LLM-realized conditional projection and
$\mathcal{R}_t$ is RuleBank.

\paragraph{Local case evidence.}
The current mini-batch indicates whether a proposed correction is overly specific to the examples that produced it. Feedback closely tied to particular entities, quantities, surface patterns, or rare edge conditions is treated as negative evidence for generalization because it is likely to introduce a narrow case patch.

\paragraph{Global recurrence evidence.}
To complement local evidence, \method{} maintains
\begin{equation}
    \mathcal{R}_t
    =
    \{(r,m_t(r)):r\in\mathcal{U}_t\},
    \label{eq:rulebank-definition}
\end{equation}
where $r$ is a canonical behavioral directive and $m_t(r)$ records how often it has matched previously accepted purified gradients. Accepted gradients are canonicalized and either matched to an existing entry or inserted as a new directive, allowing semantically equivalent corrections to accumulate recurrence evidence across surface forms.

The true scope $s(r)$ is not numerically estimated during optimization. Instead, $m_t(r)$ acts only as a \emph{soft prior}: repeatedly recovered directives receive stronger support for generality than one-off corrections, without assuming a numerical mapping from recurrence to scope or using a hard threshold.

\paragraph{Purification decision.}
Combining local specificity and recurrence evidence, $\Pi_{\text{gen}}$ classifies each gradient as a
\texttt{GENERALIZED\_RULE}, \texttt{CASE\_PATCH}, or \texttt{STYLE\_ONLY}.
Generalized corrections are retained and rewritten into concise, broadly applicable instructions, while case-specific and purely stylistic feedback is discarded. When multiple gradients are produced, all retained instructions are concatenated into $\tilde{g}_{\text{task}}$.

Thus, Stage~1 controls representational inefficiency at its source: local evidence suppresses sample-specific patches, while recurrence evidence favors stable task-level structure.

\subsection{Semantic Edit Regularization}
\label{subsec:semantic-edit-reg}

Gradient purification filters feedback before rewriting, but the LLM optimizer may still introduce verbosity, redundancy, or new case-specific instructions. Semantic Edit Regularization (SER) therefore monitors the realized transition $(p_{t-1},p_t)$ for degradation in the two structural channels of $\mathcal{I}(p)=C(p)W(p)$.

\paragraph{Capacity diagnostic.}
Capacity change is directly observable from token length:
\begin{equation}
    \rho_C(p_t)
    =
    \frac{C(p_t)-C(p_{t-1})}{C(p_{t-1})},
    \qquad
    b_C(p_t)
    =
    \mathbb{I}
    \big[
        \rho_C(p_t)>\tau_C
    \big],
    \label{eq:capacity-trigger}
\end{equation}
where $\tau_C$ prevents regularization from reacting to negligible length fluctuations.

\paragraph{Scope diagnostic.}
Scope narrowness is semantic and cannot be measured directly. SER therefore estimates only its direction of change using an LLM-based semantic diff analyzer:
\begin{equation}
    M_\Delta
    \big(
        p_{t-1},
        p_t,
        \mathcal{R}_t,
        \mathcal{G}_t
    \big)
    \rightarrow
    \big(
        \texttt{rules\_changed},
        \widehat{\operatorname{sgn}}(\Delta W)
    \big),
    \label{eq:mdelta}
\end{equation}
where \texttt{rules\_changed} records directives introduced, removed, or modified by the edit, and
$\widehat{\operatorname{sgn}}(\Delta W)\in\{+,-,0\}$ indicates whether scope narrowness appears to have increased, decreased, or remained unchanged. RuleBank provides historical context for distinguishing recurring task-level directives from isolated case patches. The scope trigger is
\begin{equation}
    b_W(p_t)
    =
    \mathbb{I}
    \left[
        \widehat{\operatorname{sgn}}(\Delta W)=+
    \right].
    \label{eq:scope-trigger}
\end{equation}

SER then forms the active regularization channels
\begin{equation}
    \mathcal{A}_t
    =
    \{C:b_C(p_t)=1\}
    \cup
    \{W:b_W(p_t)=1\},
    \label{eq:active-channels}
\end{equation}
corresponding to joint compression and generalization, compression only, generalization only, or no regularization.

The two diagnostics play complementary roles. The capacity channel provides
a precise structural signal but cannot distinguish useful expansion from
case-specific growth, whereas the scope channel captures semantic narrowing
but is necessarily estimated rather than directly measured. Combining them
allows SER to distinguish different forms of representational drift and
activate only the corresponding correction, instead of applying a uniform
compression or generalization penalty to every update.

\paragraph{Textual regularization signal.}
An LLM-realized controller $\Gamma$ converts the active channels and semantic changes into
\begin{equation}
    g_{\text{reg}}
    =
    \Gamma
    \big(
        \texttt{rules\_changed},
        \mathcal{A}_t
    \big)
    \in
    \mathcal{G}_{\text{text}}
    \cup
    \{\varnothing\}.
    \label{eq:g-reg}
\end{equation}
Capacity degradation induces compression guidance, while scope degradation induces guidance to generalize or remove narrow case patches. When neither channel is active, $g_{\text{reg}}=\varnothing$.

Thus, capacity degradation is measured exactly, whereas scope degradation is estimated only directionally. SER does not numerically evaluate $W(p)$ or $\mathcal{I}(p)$, and intervenes only when the realized edit provides evidence of structural degradation.

\begin{table*}[t]
\centering
\caption{\small{
\textbf{Cross-dataset and cross-model generalization results.} We report out-of-domain test accuracy (\%), averaged over four independent optimization seeds, and the relative improvement over TextGrad. The best and second-best results are highlighted with \textbf{bold} and \underline{underline}, respectively.
}}
\vspace{-5pt}
\label{tab:cross_dataset}
\scriptsize{
\resizebox{0.9\linewidth}{!}{
    \setlength\tabcolsep{2pt}
    \renewcommand\arraystretch{1.1}
    \begin{tabular}{c | c || c c | c c | c c}
    \toprule
    \noalign{\vskip -\belowrulesep}
    \rowcolor{CadetBlue!15}
    & & \multicolumn{2}{c|}{\textbf{Logical Ded. 3obj}} & \multicolumn{2}{c|}{\textbf{Tracking Shuf. 3obj}} & \multicolumn{2}{c}{\textbf{GSM8K}} \\
    \cline{3-8}
    \rowcolor{CadetBlue!15}
    \multirow{-2}{*}{\textbf{Test Engine}} & \multirow{-2}{*}{\textbf{Method}} 
    & \textbf{5obj} & \textbf{7obj} & \textbf{5obj} & \textbf{7obj} & \textbf{SVAMP} & \textbf{MultiArith} \\
    \hline
    \hline
    \multirow{4}{*}{\shortstack{Qwen2 \\ -7B-Instruct}}
    & CoT      & $ 51.6\greenup{0.3} $ & $ 47.4\greenup{0.8} $ & $ \underline{42.0}\greenup{3.8} $ & $ 33.9\greenup{0.8} $ & $ \underline{89.8}\greenup{0.5} $ & $ 94.7\reddown{1.1} $ \\
    & \cellcolor{gray!10} TextGrad & \cellcolor{gray!10}$ 51.3 $ & \cellcolor{gray!10}$ 46.6 $ & \cellcolor{gray!10}$ 38.2 $ & \cellcolor{gray!10}$ 33.1 $ & \cellcolor{gray!10}$ 89.3 $ & \cellcolor{gray!10}$ 95.8 $ \\
    & REVOLVE  & $ \underline{54.4}\greenup{3.1} $ & $ \underline{47.6}\greenup{1.0} $ & $ 40.3\greenup{2.1} $ & $ \mathbf{40.2\greenup{7.1}} $ & $ 89.2\reddown{0.1} $ & $ \underline{96.2}\greenup{0.4} $ \\
    & \cellcolor{gray!10}\textbf{TextReg}  & \cellcolor{gray!10}$ \mathbf{55.3\greenup{4.0}} $ & \cellcolor{gray!10}$ \mathbf{47.8\greenup{1.2}} $ & \cellcolor{gray!10}$ \mathbf{45.4\greenup{7.2}} $ & \cellcolor{gray!10}$ \underline{39.1}\greenup{6.0} $ & \cellcolor{gray!10}$ \mathbf{90.1\greenup{0.8}} $ & \cellcolor{gray!10}$ \mathbf{96.5\greenup{0.7}} $ \\
    \hline
    \multirow{4}{*}{\shortstack{Llama-3.1 \\ -8B-Instruct}}
    & CoT      & $ 59.7\reddown{0.1} $ & $ 50.4\greenup{0.0} $ & $ 58.8\reddown{10.9} $ & $ 48.5\reddown{18.2} $ & $ \mathbf{85.5\greenup{0.7}} $ & $ \underline{96.0}\greenup{0.0} $ \\
    & \cellcolor{gray!10}TextGrad & \cellcolor{gray!10}$ \underline{59.8} $ & \cellcolor{gray!10}$ 50.4 $ & \cellcolor{gray!10}$ 69.7 $ & \cellcolor{gray!10}$ \underline{66.7} $ & \cellcolor{gray!10}$ 84.8 $ & \cellcolor{gray!10}$ \underline{96.0} $ \\
    & REVOLVE  & $ 57.7\reddown{2.1} $ & $ \underline{50.6}\greenup{0.2} $ & $ \underline{76.0}\greenup{6.3} $ & $ 65.4\reddown{1.3} $ & $ 84.6\reddown{0.2} $ & $ 95.3\reddown{0.7} $ \\
    & \cellcolor{gray!10}\textbf{TextReg}  & \cellcolor{gray!10}$ \mathbf{61.1\greenup{1.3}} $ & \cellcolor{gray!10}$ \mathbf{51.0\greenup{0.6}} $ & \cellcolor{gray!10}$ \mathbf{79.7\greenup{10.0}} $ & \cellcolor{gray!10}$ \mathbf{76.6\greenup{9.9}} $ & \cellcolor{gray!10}$ \mathbf{85.5\greenup{0.7}} $ & \cellcolor{gray!10}$ \mathbf{96.7\greenup{0.7}} $ \\
    \hline
    \multirow{4}{*}{\shortstack{Llama-3 \\ -8B-Instruct}}
    & CoT      & $ \underline{52.6}\greenup{10.5} $ & $ \underline{48.8}\greenup{7.7} $ & $ 35.6\reddown{10.3} $ & $ 36.5\reddown{1.5} $ & $ 83.7\greenup{0.5} $ & $ 94.7\greenup{0.0} $ \\
    & \cellcolor{gray!10}TextGrad & \cellcolor{gray!10}$ 42.1 $ & \cellcolor{gray!10}$ 41.1 $ & \cellcolor{gray!10}$ 45.9 $ & \cellcolor{gray!10}$ 38.0 $ & \cellcolor{gray!10}$ 83.2 $ & \cellcolor{gray!10}$ 94.7 $ \\
    & REVOLVE  & $ 51.5\greenup{9.4} $ & $ \underline{48.8}\greenup{7.7} $ & $ \underline{46.1}\greenup{0.2} $ & $ \underline{41.8}\greenup{3.8} $ & $ \mathbf{83.9\greenup{0.7}} $ & $ 94.7\greenup{0.0} $ \\
    & \cellcolor{gray!10}\textbf{TextReg}  & \cellcolor{gray!10}$ \mathbf{53.3\greenup{11.2}} $ & \cellcolor{gray!10}$ \mathbf{52.0\greenup{10.9}} $ & \cellcolor{gray!10}$ \mathbf{54.3\greenup{8.4}} $ & \cellcolor{gray!10}$ \mathbf{48.3\greenup{10.3}} $ & \cellcolor{gray!10}$ \mathbf{83.9\greenup{0.7}} $ & \cellcolor{gray!10}$ \mathbf{94.9\greenup{0.2}} $ \\
    \hline
    \multirow{4}{*}{\shortstack{Phi-3.5 \\ -Mini-Instruct}}
    & CoT      & $ \underline{57.0}\greenup{10.9} $ & $ \underline{50.5}\greenup{4.1} $ & $ \underline{89.6}\greenup{6.0} $ & $ 89.1\reddown{3.7} $ & $ \mathbf{90.4\greenup{9.8}} $ & $ \mathbf{97.0\greenup{10.2}} $ \\
    & \cellcolor{gray!10}TextGrad & \cellcolor{gray!10}$ 46.1 $ & \cellcolor{gray!10}$ 46.4 $ & \cellcolor{gray!10}$ 83.6 $ & \cellcolor{gray!10}$ \mathbf{92.8} $ & \cellcolor{gray!10}$ 80.6 $ & \cellcolor{gray!10}$ 86.8 $ \\
    & REVOLVE  & $ 43.4\reddown{2.7} $ & $ 38.7\reddown{7.7} $ & $ 86.8\greenup{3.2} $ & $ 87.7\reddown{5.1} $ & $ 87.0\greenup{6.4} $ & $ \underline{95.1}\greenup{8.3} $ \\
    & \cellcolor{gray!10}\textbf{TextReg}  & \cellcolor{gray!10}$ \mathbf{57.9\greenup{11.8}} $ & \cellcolor{gray!10}$ \mathbf{55.2\greenup{8.8}} $ & \cellcolor{gray!10}$ \mathbf{94.1\greenup{10.5}} $ & \cellcolor{gray!10}$ \underline{92.2}\reddown{0.6} $ & \cellcolor{gray!10}$ \underline{88.5}\greenup{7.9} $ & \cellcolor{gray!10}$ 94.7\greenup{7.9} $ \\
    \noalign{\vskip -\belowrulesep}
    \bottomrule
    \end{tabular}
}}
\vspace{-15pt}
\end{table*}

\subsection{Regularization-Guided Prompt Update}
\label{subsec:reg-guided-update}

After the first two stages, \method{} obtains $\tilde{g}_{\text{task}}$ and, when needed, $g_{\text{reg}}$. The final stage uses regularization to shape \emph{how} the task correction is realized while keeping task fidelity primary.

This separation between task fidelity and structural preference is important:
regularization should shape the form of an update without suppressing a
correction that is necessary for task performance. Stage~3 therefore treats
$\tilde{g}_{\text{task}}$ as defining what must be changed, while
$g_{\text{reg}}$ constrains how that change should be expressed. This
asymmetry preserves the optimization objective while discouraging the
unnecessary prompt expansion and narrow rule accumulation associated with
distributional overfitting.

This design also avoids forcing every beneficial update toward shorter or more generic wording. A task-relevant instruction may legitimately increase prompt length or introduce a specialized distinction, and \method{} does not suppress such changes solely because they affect $C(p)$ or $W(p)$. Regularization instead acts as a preference over task-faithful rewrites, intervening only when a structurally more efficient realization is available.

Let $\mathcal{E}(p_t,\tilde{g}_{\text{task}})$ denote the \emph{task-faithful set}: conceptually, prompt rewrites that faithfully implement $\tilde{g}_{\text{task}}$. Within this set, regularization favors an edit most compatible with $g_{\text{reg}}$:
\begin{equation}
    p_{t+1}
    =
    \arg\min_{p'
    \in
    \mathcal{E}(p_t,\tilde{g}_{\text{task}})}
    \Phi
    \big(
        p'-p_t,\,
        g_{\text{reg}}
    \big),
    \label{eq:reg-guided-update}
\end{equation}
where $\Phi$ denotes incompatibility between the semantic edit and the active regularization guidance, and $p'-p_t$ denotes the semantic edit rather than literal vector subtraction.

\Cref{eq:reg-guided-update} is a conceptual characterization of the rewrite preference implemented by $\mathcal{U}$, not an explicitly enumerated numerical optimization: \method{} does not construct $\mathcal{E}$ or evaluate $\Phi$ as a scalar. Instead, the optimizer LLM realizes this preference directly in natural language.

This task-dominance fallback prevents regularization from becoming a hard structural constraint. Together, the three stages control which feedback enters optimization, detect structural drift after rewriting, and coordinate task improvement with structural guidance, thereby operationalizing~\cref{eq:regularized-objective} without directly evaluating $\mathcal{I}(p)$ or its gradients.

\section{Experiments}
\label{sec:exp}
We assess \method along three dimensions: \textbf{Q1}: does it outperform existing 
prompt optimization methods on out-of-distribution generalization across datasets and 
test engines (Section \ref{sec:main_results})? \textbf{Q2}: how does each component contribute 
to the overall performance (Section \ref{sec:ablation})? \textbf{Q3}: how robust is it when 
individual components are degraded or replaced with weaker variants 
(Section \ref{sec:resilience})?

\subsection{Experimental Setup}
\label{sec:exp_setup}
\paragraph{Tasks and Datasets.}
We evaluate \method on six datasets from the Big Bench Hard benchmark~\citep{suzgun2023challenging,srivastava2023beyond}—\textbf{Logical Deduction} (Three / Five / Seven Objects) and \textbf{Tracking Shuffled Objects} (Three / Five / Seven Objects), as well as three arithmetic datasets: \textbf{GSM8K}~\citep{cobbe2021training}, \textbf{SVAMP}~\citep{patel2021nlp}, and \textbf{MultiArith}~\citep{roy2015solving,koncel2016mawps}. Prompts are optimized on \textbf{Logical Deduction (Three Objects)}, \textbf{Tracking Shuffled Objects (Three Objects)}, and \textbf{GSM8K}, and evaluated for cross-dataset generalization on the remaining datasets: the harder variants \textbf{Logical Deduction} (Five / Seven Objects) and \textbf{Tracking Shuffled Objects} (Five / Seven Objects) for the BBH tasks, and \textbf{SVAMP} and \textbf{MultiArith} for the arithmetic tasks. Our primary metric is Accuracy (Acc), measured by exact match string on the final answer~\citep{yuksekgonul2024textgrad}. Further datasets and implementation details are in Appendix~\ref{app:exp_details}.

\paragraph{LLM Backends.}
Our experiments are conducted on four open-source LLM backends as test engines: Qwen2-7B-Instruct~\citep{qwen2}, Phi-3.5-Mini-Instruct~\citep{abdin2024phi}, Llama-3-8B-Instruct, and Llama-3.1-8B-Instruct~\citep{dubey2024llama}. For prompt optimization, we use Qwen2.5-7B-Instruct~\citep{qwen25} as the forward engine that executes prompts on training samples, and GPT-4o~\citep{gpt4o} as the shared backward engine that performs all LLM-driven optimization operations. This shared configuration applies uniformly to \method and all baselines, to ensure a fair and controlled comparison. The optimized prompts are then evaluated on the four test engines above.

\paragraph{Counterparts.}
We compare \method against three baselines: Zero-shot Chain-of-Thought (CoT)~\citep{kojima2022large,wei2022chain}, TextGrad~\citep{yuksekgonul2024textgrad}, and REVOLVE~\citep{zhang2024revolve}.

\paragraph{Evaluation Protocol.}
Every entry in Table~\ref{tab:cross_dataset} for TextGrad, REVOLVE, and \method{} is the mean over four optimization runs with independent seeds: each run optimizes its own prompt, and all resulting prompts are evaluated on the same test set of each dataset with greedy decoding. CoT uses the same fixed prompt evaluated under the same protocol.

\subsection{Main Results}
\label{sec:main_results}
To answer \textbf{Q1}, Table \ref{tab:cross_dataset} reports cross-dataset and cross-model 
generalization results, where prompts optimized on three source tasks are evaluated 
out-of-distribution across four downstream test engines. \method consistently achieves the 
best or second-best accuracy on nearly all (test engine, dataset) pairs, outperforming 
both TextGrad and REVOLVE in the large majority of cells. The advantage is most 
pronounced on harder variants of the source task: on Tracking Shuffled Objects, 
\method improves over TextGrad by \textbf{+10.0} (5obj) and \textbf{+9.9} (7obj) on 
Llama-3.1-8B-Instruct, and by \textbf{+8.4} (5obj) and \textbf{+10.3} (7obj) on 
Llama-3-8B-Instruct. Notably, baseline methods often degrade out-of-distribution accuracy 
below that of the unoptimized CoT prompt---on Phi-3.5-Mini-Instruct, REVOLVE underperforms CoT
on all six datasets---directly exposing the prompt overfitting problem that motivates 
our work, while \method preserves and extends the generalization benefits of CoT. This 
robustness holds across heterogeneous test engines spanning different architectures and 
instruction-tuning recipes, confirming that mitigating prompt overfitting is a 
model-agnostic property of the optimization procedure.

\subsection{Ablation Study}
\label{sec:ablation}
\begin{wrapfigure}{r}{0.5\textwidth}
    \vspace{-18pt}
    \begin{center}
        \includegraphics[width=0.48\textwidth]{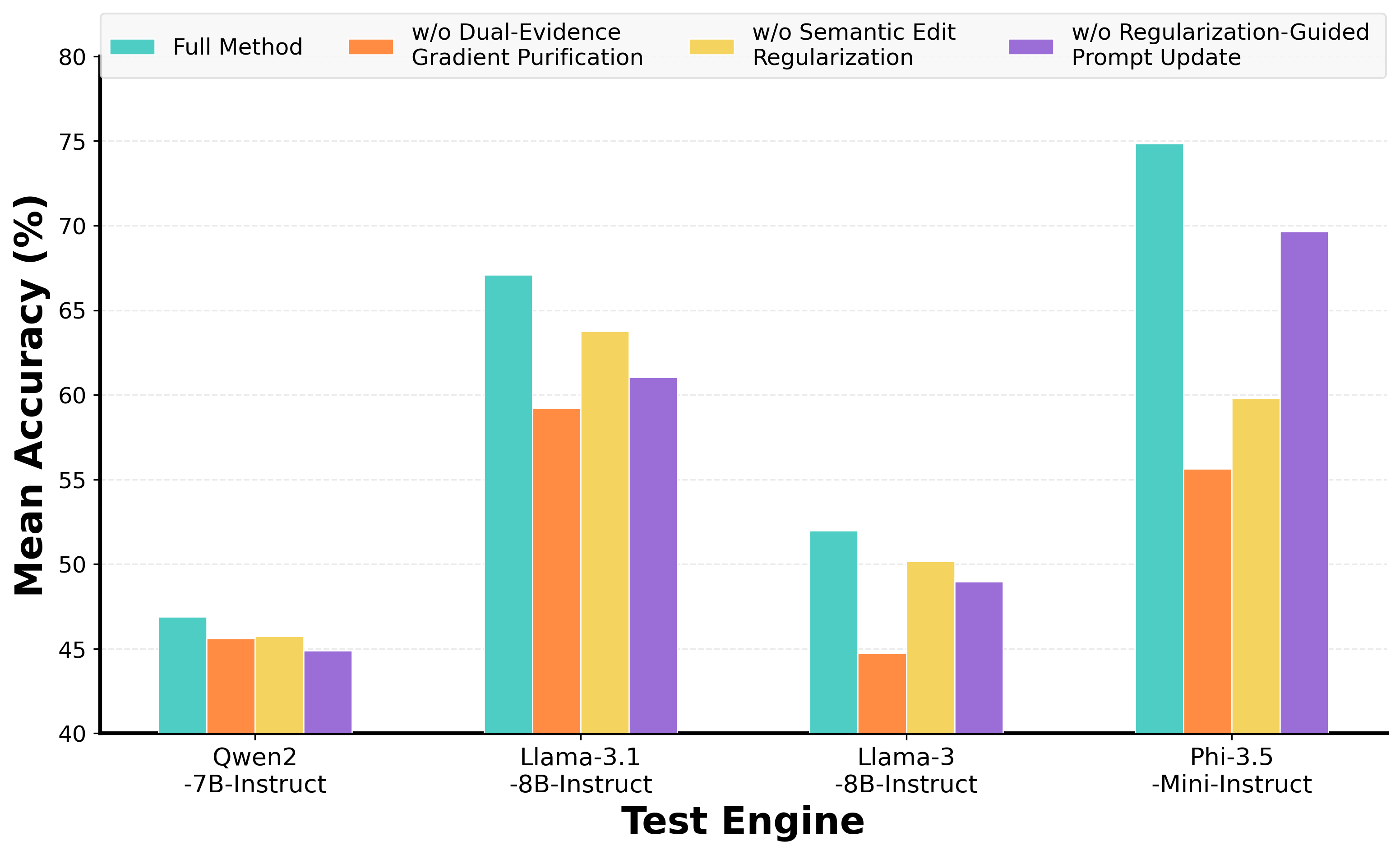}
    \end{center}
    \vspace{-10pt}
    \caption{Ablation study of the $3$ core components of \method{}: Gradient Purification, Semantic Edit Regularization, and Regularization-Guided Update. Each bar reports mean out-of-distribution accuracy (\%) across four OOD tasks (Tracking Shuffled Objects 5/7 obj, Logical Deduction 5/7 obj) on each test engine. Section \ref{sec:ablation} presents analyses.}
    \label{fig:ablation_studies}
    \vspace{-20pt}
\end{wrapfigure}
To address \textbf{Q2}, we disable each of \method{}'s three components in turn 
(\cref{fig:ablation_studies}). Each bar reports mean out-of-distribution accuracy across four 
OOD tasks---Tracking Shuffled Objects (5/7 obj) and Logical Deduction (5/7 obj)---using 
prompts optimized on the corresponding 3-object source tasks. Full \method{} achieves 
the highest accuracy on all four test engines, and removing any of \textbf{Dual-Evidence 
Gradient Purification}, \textbf{Semantic Edit Regularization}, or 
\textbf{Regularization-Guided Update} produces a clear and consistent drop. This 
confirms that the three components are jointly necessary: source-level gradient 
filtering, post-edit inefficiency diagnosis, and regularization-guided rewriting each 
contribute non-redundantly to mitigating prompt distributional overfitting.

\subsection{Resilience Study}
\label{sec:resilience}

\begin{figure*}[t]
    \vspace{-15pt}
    \centering
    \captionsetup[sub]{font=scriptsize}
    
    \begin{minipage}[b]{0.24\textwidth}
        \centering
        \includegraphics[width=\linewidth]{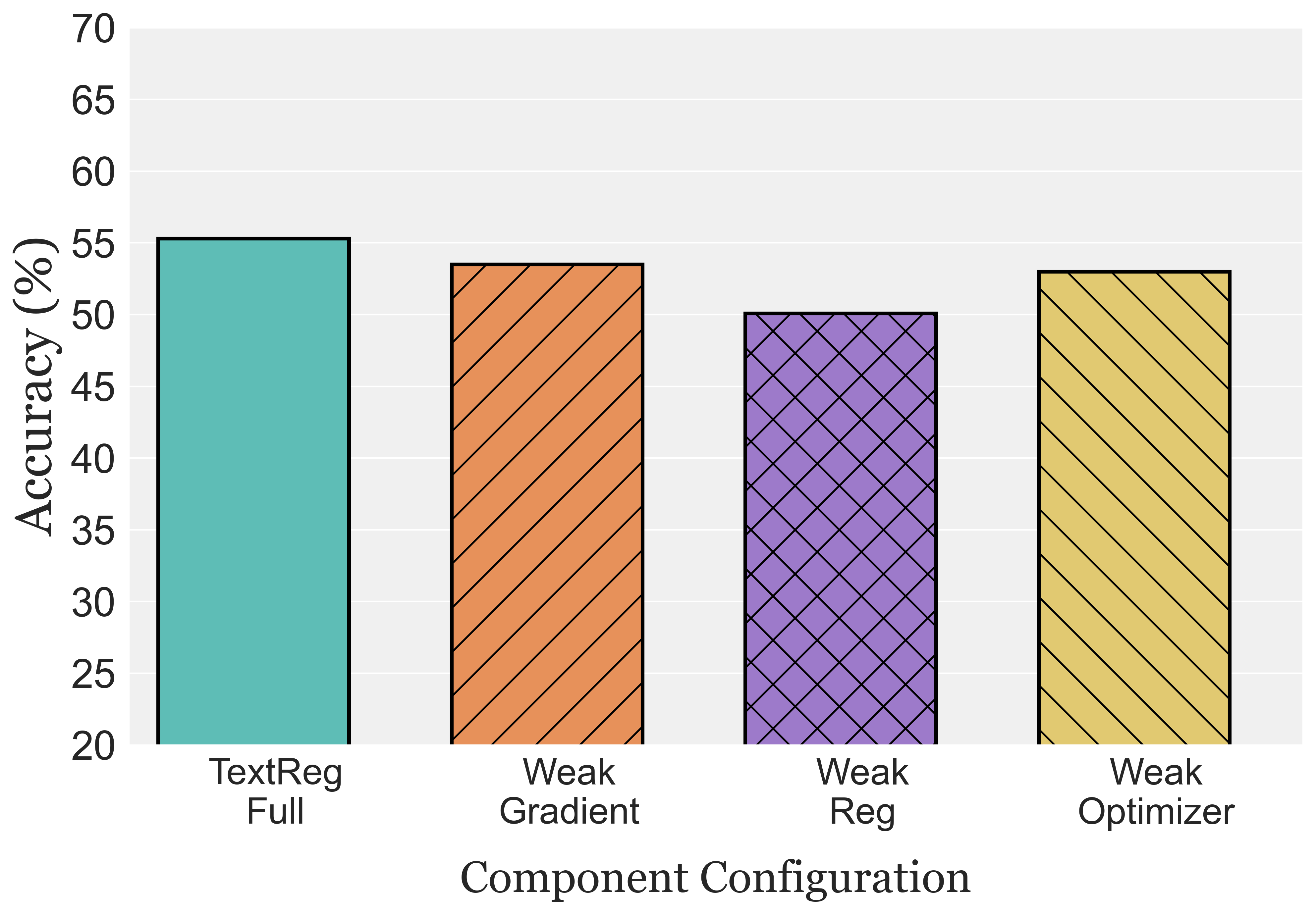}
        \subcaption{Logical Ded. 5 obj}
        \label{fig:resilience_logical_5}
    \end{minipage}
    \hfill
    \begin{minipage}[b]{0.24\textwidth}
        \centering
        \includegraphics[width=\linewidth]{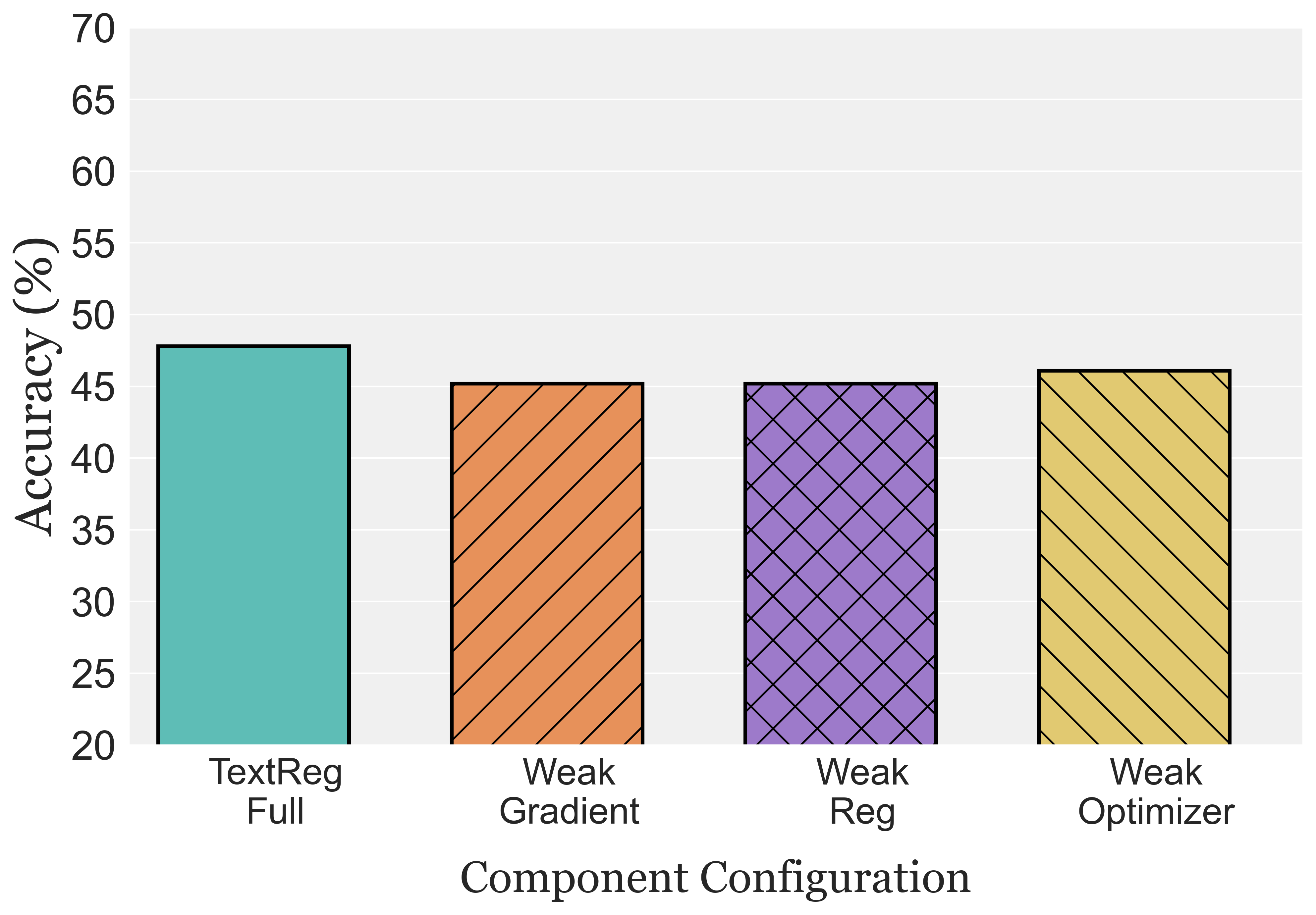}
        \subcaption{Logical Ded. 7 obj}
        \label{fig:resilience_logical_7}
    \end{minipage}
    \hfill
    \begin{minipage}[b]{0.24\textwidth}
        \centering
        \includegraphics[width=\linewidth]{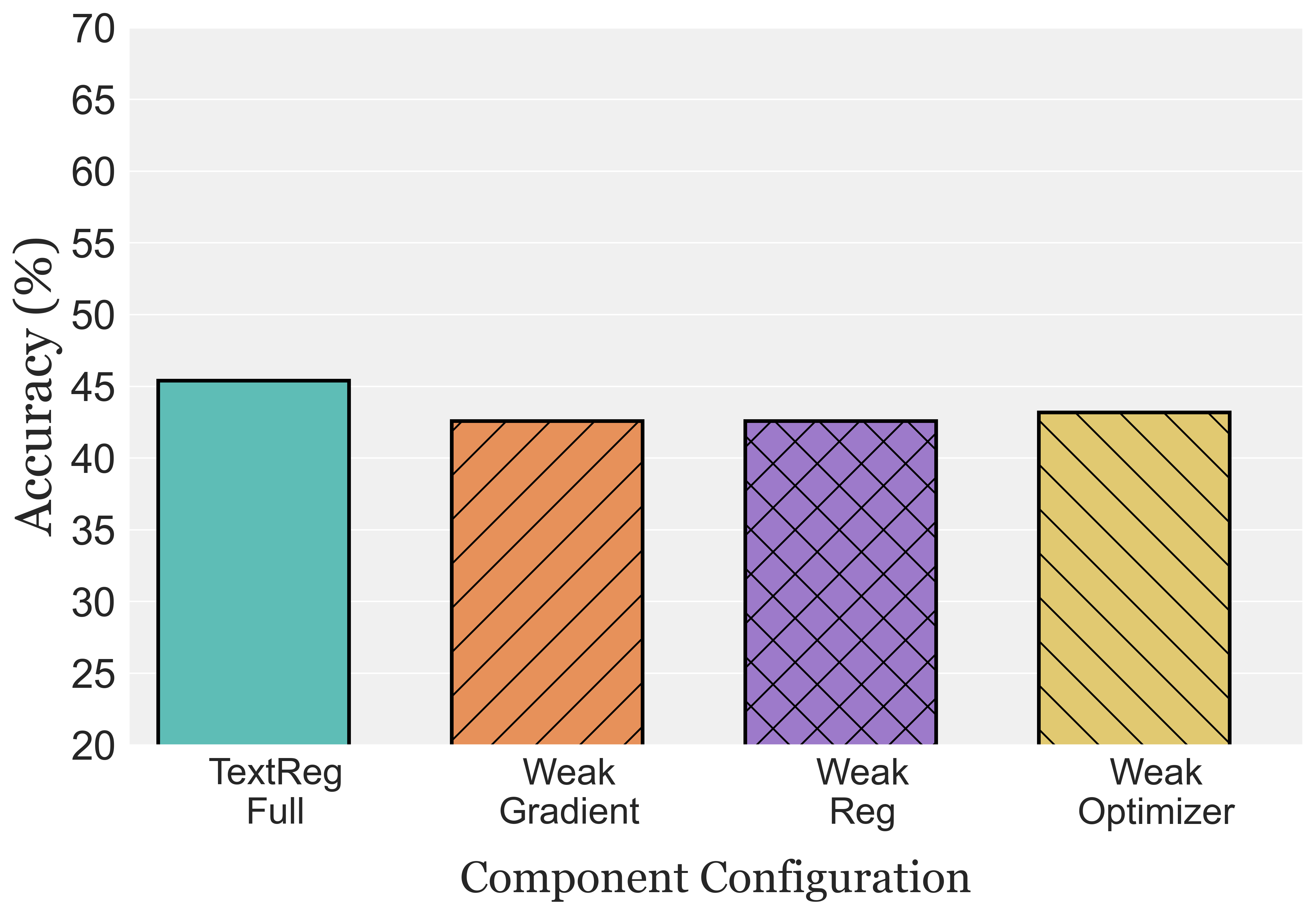}
        \subcaption{Tracking Shuf. 5obj}
        \label{fig:resilience_tracking_5}
    \end{minipage}
    \hfill
    \begin{minipage}[b]{0.24\textwidth}
        \centering
        \includegraphics[width=\linewidth]{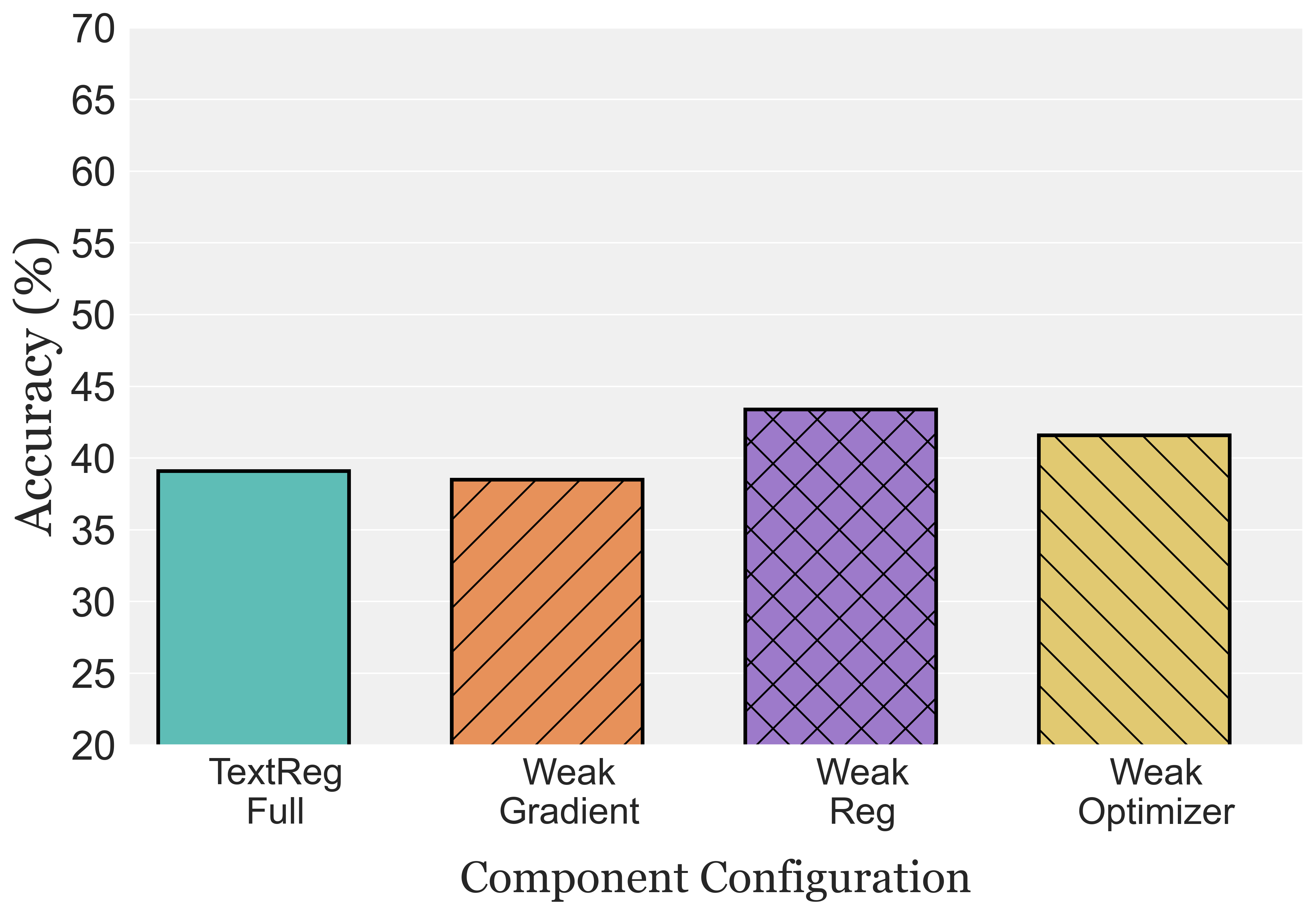}
        \subcaption{Tracking Shuf. 7obj}
        \label{fig:resilience_tracking_7}
    \end{minipage}
    
    \vspace{-1mm}
    \caption{Resilience analysis of \method{} under role-wise engine degradation, where each of the three LLM-driven roles in the optimization pipeline is replaced one at a time with a weaker Qwen2.5-7B-Instruct model. For an in-depth analysis, please refer to Section \ref{sec:resilience}.}
    \label{fig:resilience}
    \vspace{-6mm}
\end{figure*}

To probe \method{}'s resilience (\textbf{Q3}), we replace one of three LLM-driven roles 
at a time with a substantially weaker LLM: \textbf{Gradient} (textual feedback, 
gradient purification, RuleBank rule extraction), \textbf{Regularization} (semantic 
edit analysis $M_\Delta$, regularization gradient synthesis), or \textbf{Optimizer} 
(prompt rewriting from both signals). Starting from the \textbf{All-Strong} baseline 
where all three roles use GPT-4o, we downgrade each role in turn to Qwen2.5-7B-Instruct.
As shown in \cref{fig:resilience}, \method{} retains strong performance under all three 
weakenings, with only minor accuracy drops; on the hardest variant Tracking Shuffled 
Objects (7 obj), \textbf{Weak Regularization} and \textbf{Weak Optimizer} even surpass 
\textbf{All-Strong}, indicating that \method{}'s regularization signal is structural 
rather than capability-bound and that parts of the optimization loop can be served by 
lightweight models.

\section{Conclusion}
\label{sec:conclusion}
We frame prompt distributional overfitting as a failure of representational efficiency:
optimized prompts may reduce training loss by expanding in length and accumulating 
narrow, sample-specific rules, hurting OOD generalization. We formalize this through 
\emph{representational inefficiency}, the interaction between capacity cost and scope
narrowness, and propose \method, which controls its growth through Dual-Evidence Gradient
Purification, Semantic Edit Regularization, and Regularization-Guided Prompt Update. 
\method improves OOD generalization over existing methods across multiple reasoning 
benchmarks. Our evaluation focuses primarily on single-turn reasoning with
well-defined behavioral rules. Appendix~\ref{app:beyond} provides preliminary
results on open-ended generation, broader generation tasks, multi-turn prompts,
and agent instructions remain directions for future work. In these settings, rule
structure can be less explicit and capacity cost may be shared across turns.

\bibliographystyle{unsrt}
\bibliography{8_reference}

\newpage
\appendix
\section{Experimental Details}
\label{app:exp_details}

\subsection{Dataset Details}
\label{sec:exp_datasets}
This section elaborates on the datasets summarized in Section~\ref{sec:exp_setup}. We 
conduct experiments on nine reasoning datasets that span symbolic and arithmetic 
domains, deliberately chosen so that each source task has well-defined harder or 
related variants for testing cross-dataset generalization. Following common practice in 
prompt optimization~\citep{yuksekgonul2024textgrad}, we report strict string-based 
exact-match accuracy throughout. Each dataset is described below.

\begin{itemize}[leftmargin=*]
    \item \textbf{Big-Bench Hard (BBH)}~\citep{suzgun2023challenging,srivastava2023beyond}. 
    A suite of 23 challenging multi-step reasoning tasks from BIG-Bench on which prior 
language models fell short of average human performance. We draw two task families 
from BBH, each with three difficulty levels parameterized by the number of objects 
involved:
    \begin{itemize}[leftmargin=*, topsep=2pt, itemsep=0pt, partopsep=0pt]
        \item \textbf{Logical Deduction (3 / 5 / 7 Objects):} Deduce the order of a 
        sequence of objects from clues describing their spatial relationships and 
        placements, then answer a query about the inferred ordering.
        \item \textbf{Tracking Shuffled Objects (3 / 5 / 7 Objects):} Given the initial 
        positions of a set of objects and a sequence of pairwise swaps applied to them, 
        determine the final position of a specified object.
    \end{itemize}
    Prompts are optimized on the 3-object variant of each family and evaluated 
out-of-distribution on the 5- and 7-object variants, allowing us to measure how well 
optimized prompts transfer to harder instances of the same task structure. For the 
3-object variants used as source tasks, we follow TextGrad~\citep{yuksekgonul2024textgrad} 
and adopt a 50 / 100 / 100 train / validation / test split.

    \item \textbf{GSM8K}~\citep{cobbe2021training}. A widely used benchmark of 
linguistically diverse grade-school math word problems requiring multi-step 
arithmetic. We use it as the source task for arithmetic optimization.

\item \textbf{SVAMP}~\citep{patel2021nlp}. A challenge benchmark constructed by 
applying targeted variations to existing math word problems, designed to probe 
robustness rather than surface pattern matching.

\item \textbf{MultiArith}~\citep{roy2015solving,koncel2016mawps}. A set of multi-step 
arithmetic word problems requiring the composition of multiple elementary operations.
\end{itemize}

\subsection{Counterpart Details}
\label{sec:counterparts}
We summarize the three baselines compared against \method.

\begin{itemize}[leftmargin=*]
    \item \textbf{Zero-shot Chain-of-Thought (CoT)}~\citep{kojima2022large,wei2022chain}. 
    A foundational baseline that prompts the model with cues such as ``Think 
    step-by-step'' to elicit multi-step reasoning prior to producing the final answer.
    
    \item \textbf{TextGrad}~\citep{yuksekgonul2024textgrad}. A first-order optimization 
    method that uses natural-language feedback from an evaluator LLM as a ``textual 
    gradient'', iteratively refining the prompt from immediate, local feedback signals.
    
    \item \textbf{REVOLVE}~\citep{zhang2024revolve}. An optimization method that builds 
    on first-order techniques by tracking how model responses evolve across iterations. 
    Through this historical context, REVOLVE seeks more stable optimization and avoids 
    the local optima that often trap methods relying solely on single-step feedback.
\end{itemize}

\subsection{Implementation Details}
\label{sec:exp_implementation}
The experimental pipeline follows Section~\ref{sec:exp_setup} and applies uniformly 
to \method and all baselines. Prompt execution is performed by Qwen2.5-7B-Instruct~\citep{qwen25} 
as the forward engine, while GPT-4o~\citep{gpt4o} acts as the shared backward engine 
responsible for all LLM-driven optimization operations (gradient generation, 
purification, semantic edit analysis, regularization synthesis, and prompt rewriting). 
Optimized prompts are evaluated on four open-source test engines: Qwen2-7B-Instruct~\citep{qwen2}, 
Phi-3.5-Mini-Instruct~\citep{abdin2024phi}, Llama-3-8B-Instruct, and 
Llama-3.1-8B-Instruct~\citep{dubey2024llama}.

For iterative optimization, we adopt the same training budget across all methods, 
using a batch size of 3 over 12 optimization iterations (36 training samples in 
total), matching the protocol established in REVOLVE~\citep{zhang2024revolve}. For 
LLM generation, we allow a maximum of 2000 new tokens with a top-$p$ of 0.99, and 
set the decoding temperature to 0 throughout to ensure reproducibility. \method{}'s 
only hyperparameter, the relative length-growth threshold $\tau_C$ for the capacity 
channel of Semantic Edit Regularization, is set to $\tau_C = 0.2$ in all 
experiments. The only deviation from REVOLVE's protocol is on GSM8K: we relax the 
validation acceptance criterion by $1\%$ (i.e., a prompt update is accepted 
whenever validation accuracy does not drop by more than $1\%$), since under the 
strict monotone criterion the prompt almost never updates on this task; this 
relaxation is applied uniformly to all methods.

All experiments are conducted on a server with six NVIDIA A100 80GB GPUs.

Unless stated otherwise, every reported number for TextGrad, REVOLVE, and 
\method{} is the mean over four independent optimization seeds, each of which 
produces a separately optimized prompt; CoT uses the same fixed prompt evaluated 
over four runs. OOD evaluation uses the full released splits: 250 examples for 
each BBH variant, 1{,}000 for SVAMP, and 600 for MultiArith. The 100-example 
splits above are the held-out source-task test sets used for the in-distribution 
results in Appendix~\ref{app:id}.

\subsection{Initial Prompts}
\label{app:init_prompts}
All methods and all seeds start from the same fixed, task-agnostic prompt for 
each task. The initializations are inherited from the zero-shot prompts released 
with TextGrad~\citep{yuksekgonul2024textgrad} and are not tuned for any method: 
GSM8K uses the released prompt unchanged, and the two BBH tasks use the same 
wording with only the answer-format clause adapted to multiple-choice outputs. 
Appendix~\ref{app:init} studies how sensitive the results are to this choice.
\begin{tcolorbox}[
    enhanced, breakable,
    colframe=black!70, colback=yellow!5,
    boxrule=1pt, arc=4mm,
    left=2mm, right=2mm, top=1mm, bottom=1mm,
    title={Initial Prompts}
]
\small
\textbf{GSM8K} (released TextGrad prompt; the spelling of ``mathemetical'' is 
preserved from the original): 
\textit{You will answer a mathemetical reasoning question. Think step by step. 
The last line of your response should be of the following format: `Answer: 
\$VALUE' where VALUE is a numerical value.}
\\[1ex]
\textbf{Logical Deduction and Tracking Shuffled Objects}: 
\textit{You will answer a reasoning question. Think step by step. The last line 
of your response should be of the following format: `Answer: (\$VALUE)' where 
VALUE is the letter of the correct option.}
\end{tcolorbox}

\section{Attribution of the Gains: Compute Budget and Additional Optimizers}
\label{app:attribution}

The OOD gains in \cref{tab:cross_dataset} could arise from factors other than regularization, including additional backward-engine calls, limited baseline coverage, or prompt-length control alone. We examine each possibility in turn. Over a complete 12-step run, \method{} uses 174 backward calls, compared with 84 for TextGrad and REVOLVE, due to gradient purification, RuleBank canonicalization, semantic diff analysis, and regularization synthesis. This increases the GPT-4o cost from \$0.63 for TextGrad to \$0.98 for \method{} and adds approximately 13\% to the wall-clock time, while the shorter prompts reduce solver input tokens by 36\% during optimization.

\paragraph{Matched budget.}
To isolate the effect of additional compute, we rerun TextGrad and REVOLVE with a task-only best-of-$N$ search. At each step, eight candidate rewrites are sampled and one additional call selects among them using task fidelity alone, without any notion of length, scope, or generality. This yields 180 backward calls per run, slightly more than the 174 used by \method{}.

\paragraph{Additional optimizers.}
We further compare against OPRO~\citep{yang2024large}, DSPy-MIPROv2~\citep{khattab2023dspy,opsahl2024optimizing}, DLPO~\citep{peng2025dlpo}, REMO~\citep{wu2025reflection}, and a CAPO-style length penalty~\citep{zehle2025capo} attached to the TextGrad backbone. DLPO uses its official implementation, while REMO is re-implemented from the paper using its mistake notebook, top-$k=5$ retrieval, and meta-controller. For the CAPO-style control, candidate scores are penalized according to prompt length using CAPO's default coefficient $\gamma=0.05$, normalized by the maximum prompt length observed within the run. We attach this penalty to TextGrad rather than run the full population-based CAPO pipeline so that the starting prompt, search procedure, and optimization setting remain controlled, isolating the effect of constant length regularization.

\begin{table*}[t]
\centering
\caption{\small{
\textbf{Attribution of the OOD gains.} Out-of-distribution accuracy (\%) averaged over the four test engines for the default-budget baselines, matched-budget best-of-$N$ controls (180 backward calls per run vs.\ 174 for \method{}), and five additional comparison methods. The best and second-best results are highlighted with \textbf{bold} and \underline{underline}, respectively.
}}
\vspace{-5pt}
\label{tab:attribution}
\scriptsize{
\resizebox{\linewidth}{!}{
    \setlength\tabcolsep{3pt}
    \renewcommand\arraystretch{1.1}
    \begin{tabular}{c || c c | c c | c c c c c | c}
    \toprule
    \noalign{\vskip -\belowrulesep}
    \rowcolor{CadetBlue!15}
    & \multicolumn{2}{c|}{\textbf{Default budget}} & \multicolumn{2}{c|}{\textbf{Matched budget (best-of-$N$)}} & \multicolumn{5}{c|}{\textbf{Additional optimizers}} & \\
    \cline{2-11}
    \rowcolor{CadetBlue!15}
    \multirow{-2}{*}{\textbf{OOD Dataset}} & \textbf{TextGrad} & \textbf{REVOLVE} & \textbf{TextGrad} & \textbf{REVOLVE} & \textbf{OPRO} & \textbf{MIPROv2} & \textbf{DLPO} & \textbf{REMO} & \textbf{CAPO-pen.} & \multirow{-2}{*}{\textbf{TextReg}} \\
    \hline
    \hline
    Logical Ded. 5obj  & $ 49.8 $ & $ 51.8 $ & $ \underline{54.9} $ & $ 53.6 $ & $ 53.2 $ & $ \underline{54.9} $ & $ 54.8 $ & $ 53.6 $ & $ 54.8 $ & \cellcolor{gray!10}$ \mathbf{56.9} $ \\
    Logical Ded. 7obj  & $ 46.1 $ & $ 46.4 $ & $ \underline{50.1} $ & $ 47.6 $ & $ 46.9 $ & $ 46.9 $ & $ 47.4 $ & $ 48.1 $ & $ 48.5 $ & \cellcolor{gray!10}$ \mathbf{51.5} $ \\
    Tracking Shuf. 5obj & $ 59.4 $ & $ 62.3 $ & $ 63.0 $ & $ 57.9 $ & $ 59.7 $ & $ 58.6 $ & $ \underline{67.9} $ & $ 61.1 $ & $ 60.2 $ & \cellcolor{gray!10}$ \mathbf{68.4} $ \\
    Tracking Shuf. 7obj & $ 57.7 $ & $ 58.8 $ & $ \underline{63.6} $ & $ 53.0 $ & $ 54.5 $ & $ 56.1 $ & $ 61.3 $ & $ 57.2 $ & $ 58.6 $ & \cellcolor{gray!10}$ \mathbf{64.1} $ \\
    \noalign{\vskip -\belowrulesep}
    \bottomrule
    \end{tabular}
}}
\vspace{-8pt}
\end{table*}

\method{} achieves the highest mean accuracy on all four OOD evaluations in \cref{tab:attribution}. Matching the call budget improves TextGrad substantially but does not close the gap, while REVOLVE benefits only on Logical Deduction. The CAPO-style length penalty, DLPO, and REMO also remain below \method{}. Together, these controls indicate that the gains cannot be explained by additional LLM calls, limited baseline coverage, or length control alone.

\section{Prompt Growth and OOD Accuracy during Optimization}
\label{app:trajectory}

The regularization objective in \cref{eq:regularized-objective} motivates controlling prompt growth when additional capacity does not translate into transferable performance. \cref{tab:trajectory} tracks prompt length and OOD accuracy along the optimization trajectory. TextGrad grows monotonically from 41 to 351 tokens, while its OOD accuracy fluctuates and ultimately falls below that of the starting prompt. In contrast, \method{} keeps growth bounded, reaching 135--160 tokens, and improves OOD accuracy by 5.5--6.6 points at steps 3--6 while retaining a 5.0-point gain at the final step. REVOLVE produces similarly compact prompts but consistently weaker OOD accuracy, showing that prompt length alone does not explain the improvement. RuleBank is also actively reused: a run ends with 27.4 canonical rules on average, and 79\% of accepted rule mentions match an existing entry rather than creating a new one.

\begin{table*}[t]
\centering
\caption{\small{
\textbf{Prompt length and OOD accuracy along the optimization trajectory.}
Intermediate prompts at step $t$ are evaluated on the 7-object OOD variants; values are averaged over the two BBH source tasks and two test engines (Qwen2-7B-Instruct and Llama-3-8B-Instruct). For Steps 3--12, the best and second-best OOD accuracies are highlighted with \textbf{bold} and \underline{underline}, respectively; all methods share the same Step-0 accuracy.
}}
\vspace{-5pt}
\label{tab:trajectory}
\scriptsize{
\resizebox{0.7\linewidth}{!}{
    \setlength\tabcolsep{4pt}
    \renewcommand\arraystretch{1.1}
    \begin{tabular}{c | c || c c c c c}
    \toprule
    \noalign{\vskip -\belowrulesep}
    \rowcolor{CadetBlue!15}
    \textbf{Method} & \textbf{Metric} & \textbf{Step 0} & \textbf{Step 3} & \textbf{Step 6} & \textbf{Step 9} & \textbf{Step 12} \\
    \hline
    \hline
    \multirow{2}{*}{TextGrad} & Prompt tokens & $ 41 $ & $ 246 $ & $ 264 $ & $ 289 $ & $ 351 $ \\
                              & OOD Acc (\%)  & $ 42.0 $ & $ 43.0 $ & $ \underline{42.3} $ & $ 43.2 $ & $ 39.8 $ \\
    \hline
    \multirow{2}{*}{REVOLVE}  & Prompt tokens & $ 41 $ & $ 98 $ & $ 121 $ & $ 128 $ & $ 131 $ \\
                              & OOD Acc (\%)  & $ 42.0 $ & $ \underline{45.4} $ & $ 42.2 $ & $ \underline{44.0} $ & $ \underline{43.9} $ \\
    \hline
    \multirow{2}{*}{\textbf{TextReg}} & \cellcolor{gray!10}Prompt tokens & \cellcolor{gray!10}$ 41 $ & \cellcolor{gray!10}$ 135 $ & \cellcolor{gray!10}$ 139 $ & \cellcolor{gray!10}$ 152 $ & \cellcolor{gray!10}$ 160 $ \\
                              & \cellcolor{gray!10}OOD Acc (\%)  & \cellcolor{gray!10}$ 42.0 $ & \cellcolor{gray!10}$ \mathbf{47.5} $ & \cellcolor{gray!10}$ \mathbf{48.6} $ & \cellcolor{gray!10}$ \mathbf{48.1} $ & \cellcolor{gray!10}$ \mathbf{47.0} $ \\
    \noalign{\vskip -\belowrulesep}
    \bottomrule
    \end{tabular}
}}
\vspace{-8pt}
\end{table*}

\section{In-Distribution Performance and the ID--OOD Trade-off}
\label{app:id}

Regularization can trade some in-distribution (ID) fit for OOD robustness. \cref{tab:id} reports held-out source-task accuracy using Qwen2.5-7B-Instruct, the optimization forward engine, as the test engine. On the two BBH source tasks, \method{} is on average about two points below TextGrad in ID accuracy, while these tasks also show its largest OOD improvements: +7.1/+5.4 points on Logical Deduction 5/7 objects and +9.0/+6.4 on Tracking Shuffled Objects 5/7 objects, averaged over the four test engines in \cref{tab:cross_dataset}. On GSM8K, \method{} incurs no ID loss. Moreover, all three optimizers improve over the starting prompt on the BBH source tasks (91.0 on Logical Deduction and 84.5 on Tracking), showing that \method{} does not simply revert to the seed prompt. The component-wise ID effects are consistent with the OOD ablations in \cref{fig:ablation_studies}: removing Semantic Edit Regularization raises ID accuracy from 88.8 to 90.1 but weakens OOD robustness, whereas removing Gradient Purification lowers ID accuracy to 85.8 and also harms OOD performance.

\begin{table*}[t]
\centering
\caption{\small{
\textbf{In-distribution accuracy (\%) on the source tasks.} Held-out test split of each source task, with Qwen2.5-7B-Instruct as the test engine. The best and second-best results per column are highlighted with \textbf{bold} and \underline{underline}, respectively.
}}
\vspace{-5pt}
\label{tab:id}
\scriptsize{
\resizebox{0.6\linewidth}{!}{
    \setlength\tabcolsep{5pt}
    \renewcommand\arraystretch{1.1}
    \begin{tabular}{c || c c c}
    \toprule
    \noalign{\vskip -\belowrulesep}
    \rowcolor{CadetBlue!15}
    \textbf{Method} & \textbf{Logical Ded. 3obj} & \textbf{Tracking Shuf. 3obj} & \textbf{GSM8K} \\
    \hline
    \hline
    TextGrad & $ \mathbf{92.2} $ & $ \mathbf{89.8} $ & $ \underline{90.9} $ \\
    REVOLVE  & $ \underline{91.8} $ & $ \underline{86.0} $ & $ \underline{90.9} $ \\
    \cellcolor{gray!10}\textbf{TextReg} & \cellcolor{gray!10}$ 91.5 $ & \cellcolor{gray!10}$ \underline{86.0} $ & \cellcolor{gray!10}$ \mathbf{91.5} $ \\
    \noalign{\vskip -\belowrulesep}
    \bottomrule
    \end{tabular}
}}
\vspace{-8pt}
\end{table*}

\section{Robustness of the Gains}
\label{app:robustness}

We next test whether the main findings remain robust to the backward engine, prompt initialization, test engine, and protocol choices.

\subsection{Fully Weak Backward Pipeline}
\label{app:allweak}

While \cref{sec:resilience} weakens one LLM-driven role at a time, here all three roles---gradient processing, semantic edit regularization, and prompt rewriting---use Qwen2.5-7B-Instruct simultaneously. TextGrad and REVOLVE use the same weak backward engine under the same protocol. Across the eight BBH cells on Qwen2-7B-Instruct and Llama-3-8B-Instruct, All-Weak \method{} retains 91--102\% of its All-Strong accuracy from \cref{tab:cross_dataset}. In the four matched comparisons shown in \cref{tab:allweak}, it ranks first in three and remains within one point of the best in the fourth, while producing substantially shorter prompts. These results suggest that the observed regularization effect does not depend on GPT-4o alone.

\begin{table*}[t]
\centering
\caption{\small{
\textbf{Comparison under the All-Weak pipeline.} OOD accuracy (\%) when every LLM-driven role of every optimizer uses Qwen2.5-7B-Instruct, together with the range of final prompt lengths. For accuracy, the best and second-best results are highlighted with \textbf{bold} and \underline{underline}; the shortest prompt range is shown in \textbf{bold}.
}}
\vspace{-5pt}
\label{tab:allweak}
\scriptsize{
\resizebox{0.6\linewidth}{!}{
    \setlength\tabcolsep{4pt}
    \renewcommand\arraystretch{1.1}
    \begin{tabular}{c | c || c c c}
    \toprule
    \noalign{\vskip -\belowrulesep}
    \rowcolor{CadetBlue!15}
    \textbf{Test Engine} & \textbf{OOD Dataset} & \textbf{TextGrad} & \textbf{REVOLVE} & \textbf{TextReg} \\
    \hline
    \hline
    \multirow{2}{*}{\shortstack{Qwen2 \\ -7B-Instruct}}
    & Logical Ded. 5obj   & $ 52.6 $ & $ \underline{52.8} $ & \cellcolor{gray!10}$ \mathbf{53.7} $ \\
    & Tracking Shuf. 5obj & $ \mathbf{46.6} $ & $ 45.3 $ & \cellcolor{gray!10}$ \underline{45.8} $ \\
    \hline
    \multirow{2}{*}{\shortstack{Llama-3 \\ -8B-Instruct}}
    & Logical Ded. 5obj   & $ \underline{51.8} $ & $ 46.5 $ & \cellcolor{gray!10}$ \mathbf{52.6} $ \\
    & Tracking Shuf. 5obj & $ 49.4 $ & $ \underline{50.0} $ & \cellcolor{gray!10}$ \mathbf{52.1} $ \\
    \hline
    \multicolumn{2}{c||}{Final prompt tokens} & $ 448\text{--}572 $ & $ 146\text{--}189 $ & \cellcolor{gray!10}$ \mathbf{57\text{--}84} $ \\
    \noalign{\vskip -\belowrulesep}
    \bottomrule
    \end{tabular}
}}
\vspace{-8pt}
\end{table*}

\subsection{Dependence on the Initial Prompt}
\label{app:init}

The default 41-token seed in Appendix~\ref{app:init_prompts} is short and generic, which could favor a method that regularizes toward compact prompts. We therefore re-optimize Tracking Shuffled Objects (3 objects) from two richer, task-agnostic initializations and evaluate on the 5- and 7-object variants. As shown in \cref{tab:init}, \method{} achieves the highest mean in seven of eight cells. TextGrad also continues to expand substantially from these richer seeds: starting from 55 and 57 words, its final prompts average 144 and 314 words, whereas \method{} ends between 104 and 140 words.

\begin{tcolorbox}[
    enhanced, breakable,
    colframe=black!70, colback=yellow!5,
    boxrule=1pt, arc=4mm,
    left=2mm, right=2mm, top=1mm, bottom=1mm,
    title={Alternative Initializations}
]
\small
\textbf{Hand-written}: 
\textit{Answer the following multiple-choice reasoning question. Think through 
the problem step by step and base your answer only on the information given in 
the question. Do not skip steps in your reasoning. Conclude with the final 
answer on its own line in the format `Answer: (\$VALUE)', where VALUE is the 
letter of the correct option.}
\\[1ex]
\textbf{Structured}: 
\textit{You are an assistant for multiple-choice reasoning questions. 
Task: solve the question below. 
Method: reason step by step, using only the information given in the question. 
Check: verify your conclusion against the question before answering. 
Output: end your response with a line in exactly this format: `Answer: 
(\$VALUE)' where VALUE is the letter of the correct option.}
\end{tcolorbox}

\begin{table*}[t]
\centering
\caption{\small{
\textbf{OOD accuracy (\%) under alternative initial prompts.} Prompts are optimized on Tracking Shuffled Objects 3obj from the hand-written and structured initializations above. The best and second-best results are highlighted with \textbf{bold} and \underline{underline}, respectively.
}}
\vspace{-5pt}
\label{tab:init}
\scriptsize{
\resizebox{0.7\linewidth}{!}{
    \setlength\tabcolsep{4pt}
    \renewcommand\arraystretch{1.1}
    \begin{tabular}{c | c | c || c c c}
    \toprule
    \noalign{\vskip -\belowrulesep}
    \rowcolor{CadetBlue!15}
    \textbf{Test Engine} & \textbf{Initialization} & \textbf{OOD Dataset} & \textbf{TextGrad} & \textbf{REVOLVE} & \textbf{TextReg} \\
    \hline
    \hline
    \multirow{4}{*}{\shortstack{Qwen2 \\ -7B-Instruct}}
    & \multirow{2}{*}{Hand-written} & Tracking Shuf. 5obj & $ \underline{45.0} $ & $ 42.2 $ & \cellcolor{gray!10}$ \mathbf{47.8} $ \\
    &                               & Tracking Shuf. 7obj & $ 33.4 $ & $ \underline{35.2} $ & \cellcolor{gray!10}$ \mathbf{35.4} $ \\
    \cline{2-6}
    & \multirow{2}{*}{Structured}   & Tracking Shuf. 5obj & $ \mathbf{44.6} $ & $ 43.2 $ & \cellcolor{gray!10}$ \underline{44.0} $ \\
    &                               & Tracking Shuf. 7obj & $ \underline{41.4} $ & $ 38.4 $ & \cellcolor{gray!10}$ \mathbf{41.6} $ \\
    \hline
    \multirow{4}{*}{\shortstack{Llama-3 \\ -8B-Instruct}}
    & \multirow{2}{*}{Hand-written} & Tracking Shuf. 5obj & $ 43.8 $ & $ \underline{48.6} $ & \cellcolor{gray!10}$ \mathbf{54.8} $ \\
    &                               & Tracking Shuf. 7obj & $ 40.8 $ & $ \underline{41.0} $ & \cellcolor{gray!10}$ \mathbf{44.6} $ \\
    \cline{2-6}
    & \multirow{2}{*}{Structured}   & Tracking Shuf. 5obj & $ 45.6 $ & $ \underline{45.8} $ & \cellcolor{gray!10}$ \mathbf{50.8} $ \\
    &                               & Tracking Shuf. 7obj & $ 41.4 $ & $ \underline{42.6} $ & \cellcolor{gray!10}$ \mathbf{47.0} $ \\
    \noalign{\vskip -\belowrulesep}
    \bottomrule
    \end{tabular}
}}
\vspace{-8pt}
\end{table*}

\subsection{Test Engines and Protocol Sensitivity}
\label{app:sensitivity}

\paragraph{Stronger and aligned test engines.}
Evaluating the main-run prompts on Qwen2.5-32B-Instruct~\citep{qwen25}, we find that \method{} remains best on all four BBH transfers: 91.6/88.0 on Logical Deduction 5/7 objects and 99.2/96.5 on Tracking Shuffled Objects 5/7 objects, compared with 90.6/85.8 and 97.7/95.0 for TextGrad and 90.1/85.3 and 97.7/94.9 for REVOLVE. Tracking is close to ceiling for all methods; the clearest margin appears on the less-saturated Logical Deduction 7-object setting, where \method{} leads by more than two points. The ranking also persists when the forward and test engines are aligned: on Tracking Shuffled Objects 7 objects, \method{} reaches 50.8 with Llama-3-8B-Instruct (47.6 for TextGrad, 46.6 for REVOLVE) and 93.3 with Phi-3.5-Mini-Instruct (91.6 and 90.9).

\paragraph{Capacity threshold.}
$\tau_C$ is the only numeric hyperparameter of \method{}: RuleBank insertion and matching are LLM judgments, while recurrence enters purification as a soft prior rather than a hard threshold. Sweeping $\tau_C$ over $\{0.05, 0.1, 0.2, 0.3, 0.4, 0.5\}$ on the two BBH source tasks changes 7-object OOD accuracy, averaged over the four test engines, by at most 2.5 points (54.6--57.1), while final prompt length remains between 152 and 181 tokens. This suggests that performance is not highly sensitive to the exact threshold.

\paragraph{GSM8K acceptance tolerance.}
The 1\% validation-acceptance tolerance on GSM8K (Appendix~\ref{sec:exp_implementation}) is applied identically to all methods. Under strict non-degradation, each method accepts only 0--3 of the 12 updates, leaving the resulting prompts close to initialization. The corresponding SVAMP/MultiArith accuracies, averaged over the four test engines, are 87.2/97.1 for \method{}, 86.9/96.7 for REVOLVE, and 86.8/96.4 for TextGrad. Thus, the ranking is unchanged, indicating that the GSM8K-transfer results in \cref{tab:cross_dataset} are not driven by the 1\% tolerance.

\section{Beyond Symbolic and Arithmetic Reasoning}
\label{app:beyond}

To test whether the observed behavior is specific to symbolic and arithmetic reasoning, we extend the same optimization protocol to GPQA-main~\citep{rein2023gpqa}, a knowledge-intensive science QA benchmark, and CommonGen~\citep{lin2020commongen}, a concept-to-sentence generation task. We use GPT-4o as the backward engine and Qwen2.5-7B-Instruct as both solver and test engine for 12 optimization steps. GPQA uses a 50/100/298 train/validation/test split and exact-match accuracy on the answer letter; CommonGen outputs are judged by GPT-4o for complete concept coverage and fluency. As shown in \cref{tab:beyond}, \method{} achieves the highest GPQA accuracy with less prompt growth than TextGrad. On CommonGen, TextGrad and REVOLVE both fall below the starting prompt, whereas \method{} improves over it while producing the shortest final prompt. These results provide preliminary evidence that the phenomenon and its mitigation extend beyond the original reasoning benchmarks.

\begin{table*}[t]
\centering
\caption{\small{
\textbf{Knowledge-intensive QA and open-ended generation.} Test performance (\%) after optimization and final prompt word count. GPQA-main is scored by exact match of the answer letter; CommonGen reports the percentage of outputs judged by GPT-4o to cover all concepts fluently. For task performance, the best and second-best optimizer results are highlighted with \textbf{bold} and \underline{underline}, respectively.
}}
\vspace{-5pt}
\label{tab:beyond}
\scriptsize{
\resizebox{0.7\linewidth}{!}{
    \setlength\tabcolsep{4pt}
    \renewcommand\arraystretch{1.1}
    \begin{tabular}{c | l || c | c c c}
    \toprule
    \noalign{\vskip -\belowrulesep}
    \rowcolor{CadetBlue!15}
    \textbf{Task} & \textbf{Metric} & \textbf{Starting Prompt} & \textbf{TextGrad} & \textbf{REVOLVE} & \textbf{TextReg} \\
    \hline
    \hline
    \multirow{2}{*}{GPQA-main}
    & Accuracy (\%)      & $ 30.6 $ & $ 32.0 $ & $ \underline{33.2} $ & \cellcolor{gray!10}$ \mathbf{33.9} $ \\
    & Final prompt words & $ 33 $ & $ 193 $ & $ 158 $ & \cellcolor{gray!10}$ 158 $ \\
    \hline
    \multirow{2}{*}{CommonGen}
    & Judged acc.\ (\%)  & $ 87.0 $ & $ 83.0 $ & $ \underline{84.0} $ & \cellcolor{gray!10}$ \mathbf{88.0} $ \\
    & Final prompt words & $ 31 $ & $ 93 $ & $ 54 $ & \cellcolor{gray!10}$ 48 $ \\
    \noalign{\vskip -\belowrulesep}
    \bottomrule
    \end{tabular}
}}
\vspace{-8pt}
\end{table*}
\section{Prompt Details}
\label{app:prompts}

This section presents the LLM-driven prompt templates used by \method's three 
stages: gradient purification (Stage 1), semantic edit regularization (Stage 2), 
and regularization-guided prompt update (Stage 3).

\subsection{Dual-Evidence Gradient Purification ($\Pi_{\text{gen}}$)}
This prompt drives the LLM realizing $\Pi_{\text{gen}}$. It classifies each raw 
textual gradient as a generalizable rule, narrow case patch, or purely stylistic
edit, and synthesizes the retained gradient into a concise behavioral principle. 
The RuleBank is exposed to the LLM as a historical recurrence prior.
\begin{tcolorbox}[
    enhanced, breakable,
    colframe=black!70, colback=yellow!5,
    boxrule=1pt, arc=4mm,
    left=2mm, right=2mm, top=1mm, bottom=1mm,
    title={Dual-Evidence Gradient Purification Prompt}
]
\small
You are the ``Gradient Purifier''. Your job is to decide whether a proposed 
feedback (gradient) contains genuinely generalizable improvements, and if so, 
synthesize them into a concise principle. You output either a purified summary 
or an empty string.

\paragraph{Input data}
\begin{itemize}[leftmargin=*]
    \item \texttt{<CURRENT\_SYSTEM\_PROMPT>} \texttt{\{current\_prompt\}}
    \item \texttt{<EXECUTION\_CONTEXT>}: the user input and model output that 
    triggered this gradient, \texttt{\{gradient\_context\}}
    \item \texttt{<PROPOSED\_GRADIENT>} \texttt{\{gradient\_text\}}
    \item \texttt{<RULEBANK\_SUMMARY>}: previously accepted generalizable rules 
    and their frequency, \texttt{\{rulebank\_summary\}}
\end{itemize}

\paragraph{Task}
Internally classify the gradient into one of three categories:

\textbf{Category 1 — Generalizable Logic Fix} $\to$ output purified text. 
The feedback identifies a reasoning flaw or logical gap that applies broadly 
across many inputs. If the RuleBank contains a similar rule with high 
\texttt{mention\_count}, lean toward accepting. Strip specific entities, numbers, 
and scenario details; merge overlapping points; each sentence must describe a 
distinct actionable rule.

\textbf{Category 2 — Narrow Edge-Case Patch} $\to$ output empty string. 
The fix is tailored to the rare scenario in \texttt{<EXECUTION\_CONTEXT>} and 
lacks RuleBank support.

\textbf{Category 3 — Pure Style / Formatting} $\to$ output empty string. 
The feedback only concerns tone, formatting, or presentation with zero impact 
on task correctness.

\paragraph{Output format}
Respond with valid JSON: \texttt{\{"purified\_gradient": "..."\}}.
\end{tcolorbox}

\subsection{RuleBank Canonicalization}
This prompt extracts canonical mid-level rules from each accepted purified 
gradient and either increments an existing RuleBank entry or inserts a new one, 
maintaining the \texttt{mention\_count} statistics that serve as the empirical 
proxy $\widehat{s}_t(r)$.
\begin{tcolorbox}[
    enhanced, breakable,
    colframe=black!70, colback=yellow!5,
    boxrule=1pt, arc=4mm,
    left=2mm, right=2mm, top=1mm, bottom=1mm,
    title={RuleBank Canonicalization Prompt}
]
\small
You are a rule canonicalization and matching engine. Given a raw textual 
gradient and the current RuleBank, perform two tasks:

\begin{enumerate}[leftmargin=*]
    \item Extract mid-level canonical \texttt{\{rule\_scope\}} rules from the 
    raw gradient. Remove references to specific entities, exact numbers, or 
    particular examples. Preserve structural \texttt{\{rule\_patterns\}}. Keep 
    rules at mid-level abstraction.
    \item For each extracted rule, compare it with the existing RuleBank. If 
    semantically equivalent to an existing rule (same structural pattern, not 
    just similar wording), output an \texttt{INCREMENT} operation with that 
    rule's ID. Otherwise, output an \texttt{INSERT} operation with the canonical 
    description.
\end{enumerate}

\paragraph{Input}
\texttt{[CURRENT RULEBANK]} \texttt{\{rulebank\_summary\}}; 
\texttt{[RAW GRADIENT]} \texttt{\{raw\_gradient\}}.

\paragraph{Output format}
Strictly valid JSON: \texttt{\{"operations": [\{"type": "increment", "rule\_id": 
"R3", "value": 1\}, \{"type": "insert", "canonical\_description": "...", 
"value": 1\}]\}}.
\end{tcolorbox}

\subsection{Semantic Diff Analyzer ($M_\Delta$)}
This prompt drives the LLM realizing $M_\Delta$. It compares the previous and 
current prompts at the rule level, classifies each change, and outputs an 
overall specificity direction that determines which entries appear in 
$\mathcal{A}_t$. The prompt defaults to \texttt{CASE\_PATCH} unless strong 
RuleBank or cross-context support is present.
\begin{tcolorbox}[
    enhanced, breakable,
    colframe=black!70, colback=yellow!5,
    boxrule=1pt, arc=4mm,
    left=2mm, right=2mm, top=1mm, bottom=1mm,
    title={Semantic Diff Analyzer Prompt}
]
\small
You are the ``Semantic Delta Analyzer''. Compare \texttt{PREVIOUS\_PROMPT} and 
\texttt{CURRENT\_PROMPT} at the level of behavioral rules, classify each change, 
and judge the overall specificity shift.

\paragraph{What you must not do}
Do not judge whether the task logic or new rules are correct; do not act as a 
prompt quality evaluator; do not produce character-level or word-level diffs.

\paragraph{What you must do}
Identify each rule-level change between \texttt{PREVIOUS\_PROMPT} and 
\texttt{CURRENT\_PROMPT}. Default to \texttt{CASE\_PATCH} unless there is clear 
positive evidence for \texttt{GENERALIZED\_RULE}.

\textbf{\texttt{GENERALIZED\_RULE}} — A broadly applicable, task-agnostic 
behavioral principle. Mark only if the rule is not tied to specific entities, 
numbers, or surface strings, and has at least one strong support signal: a 
semantically similar high-frequency RuleBank entry, or relevance across multiple 
execution contexts.

\textbf{\texttt{CASE\_PATCH}} — The default. Best explained as triggered by a 
single specific example in \texttt{GRADIENT\_CONTEXTS}, with no clear evidence 
it would apply beyond that example, and weak RuleBank support.

\textbf{\texttt{STYLE\_ONLY}} — Pure wording or formatting change with no 
behavioral impact.

After classifying, judge the overall specificity direction: \texttt{increase} 
(adds or strengthens patch-like rules), \texttt{decrease} (removes 
\texttt{CASE\_PATCH}s or replaces them with \texttt{GENERALIZED\_RULE}s without 
adding new ones), or \texttt{neutral} (no meaningful net change).

\paragraph{Inputs}
\texttt{<INITIAL\_PROMPT>}, \texttt{<PREVIOUS\_PROMPT>}, 
\texttt{<CURRENT\_PROMPT>}, \texttt{<RULEBANK\_SUMMARY>}, 
\texttt{<GRADIENT\_CONTEXTS>}.

\paragraph{Output format}
Strictly JSON: \texttt{\{"rules\_changed": [\{"description": "...", "type": 
"..."\}], "specificity\_direction": "..."\}}.
\end{tcolorbox}

\subsection{Regularization Gradient Generator ($\Gamma$)}
This prompt drives the LLM realizing $\Gamma$. It translates the active 
regularization directions $\mathcal{A}_t$ into concrete structural directives 
referencing specific rules in the current prompt. The prompt routes among four 
modes corresponding to elements of $\mathcal{A}_t$: 
\texttt{STRONG\_REGULARIZATION} ($\{C, W\}$), 
\texttt{COMPRESSION\_ONLY} ($\{C\}$), \texttt{GENERALIZE\_ONLY} ($\{W\}$), or 
\texttt{NO\_REGULARIZATION} ($\mathcal{A}_t = \varnothing$, in which case the 
prompt is skipped).
\begin{tcolorbox}[
    enhanced, breakable,
    colframe=black!70, colback=yellow!5,
    boxrule=1pt, arc=4mm,
    left=2mm, right=2mm, top=1mm, bottom=1mm,
    title={Regularization Gradient Generator Prompt}
]
\small
You are a structural regularization controller. You are given a 
\texttt{REGULARIZATION\_MODE}, the current prompt, and a list of recent rule 
changes. Generate precise structural regularization guidance strictly according 
to the specified mode.

\paragraph{Mode definitions}

\textbf{\texttt{STRONG\_REGULARIZATION}} — Both prompt length and rule 
specificity have increased. You must give firm directives to both compress and 
generalize: (1) merge redundant sentences expressing the same rule into a single 
shorter statement; (2) tighten verbose phrasing; (3) rewrite narrow rules as 
broader principles; (4) remove case-specific patches that cannot be generalized 
(but never remove broadly useful rules).

\textbf{\texttt{COMPRESSION\_ONLY}} — Length increased but specificity did not 
worsen. Focus only on reducing length: merge redundant sentences and tighten 
verbose phrasing while preserving behavioral intent.

\textbf{\texttt{GENERALIZE\_ONLY}} — Length is stable but new narrow rules were 
added. Focus only on broadening scope: identify narrow rules and rewrite them as 
broader principles.

\paragraph{Inputs}
\texttt{<REGULARIZATION\_MODE>}, \texttt{<CURRENT\_PROMPT>}, 
\texttt{<NEWLY\_CHANGED\_RULES>}.

\paragraph{Constraints}
Do not alter task semantics or remove rules clearly beneficial for task 
correctness. Reference specific sentences or rules from 
\texttt{<CURRENT\_PROMPT>} and \texttt{<NEWLY\_CHANGED\_RULES>} — do not give 
generic advice.

\paragraph{Output format}
Strictly valid JSON: \texttt{\{"guidance": "..."\}}.
\end{tcolorbox}

\subsection{Regularization-Guided Prompt Update — System Prompt}
This is the system message sent to the optimizer LLM at every prompt rewriting 
step.
\begin{tcolorbox}[
    enhanced, breakable,
    colframe=black!70, colback=yellow!5,
    boxrule=1pt, arc=4mm,
    left=2mm, right=2mm, top=1mm, bottom=1mm,
    title={Regularization-Guided Prompt Update System Prompt}
]
\small
You are part of an optimization system that improves text (i.e., variable). You 
will be asked to creatively and critically improve prompts, solutions to 
problems, code, or any other text-based variable. You will receive some 
feedback, and use the feedback to improve the variable. The feedback may be 
noisy; identify what is important and what is correct. Pay attention to the role 
description of the variable, and the context in which it is used. You MUST give 
your response by sending the improved variable between 
\texttt{\{new\_variable\_start\_tag\}} \texttt{\{\{improved variable\}\}} 
\texttt{\{new\_variable\_end\_tag\}} tags.

\paragraph{Tag glossary}
\texttt{<LM\_SYSTEM\_PROMPT>}, \texttt{<LM\_INPUT>}, \texttt{<LM\_OUTPUT>}, 
\texttt{<FEEDBACK>}, \texttt{<CONVERSATION>}, \texttt{<FOCUS>}, \texttt{<ROLE>}, 
\texttt{<REG\_FEEDBACK>} (\method-specific: semantic edit regularization 
feedback; follow its structural directives to control prompt length and 
specificity).
\end{tcolorbox}

\subsection{Regularization-Guided Prompt Update — User Message Trailing}
This trailing instruction is appended to the optimizer's user message, 
implementing the task-faithful selection of ~\cref{eq:reg-guided-update}: 
when task and regularization feedback target the same region, merge them into one
edit; when they target different regions, apply both; when they conflict, fold 
the task fix into the reg-aware shape, and only as a last resort prioritize the 
task item over a specific reg item.
\begin{tcolorbox}[
    enhanced, breakable,
    colframe=black!70, colback=yellow!5,
    boxrule=1pt, arc=4mm,
    left=2mm, right=2mm, top=1mm, bottom=1mm,
    title={Regularization-Guided Prompt Update User Message Trailing}
]
\small
Improve the variable (\texttt{\{variable\_desc\}}) by integrating both the task 
feedback in \texttt{<FEEDBACK>} and the regularization feedback in 
\texttt{<REG\_FEEDBACK>}. Both contain concrete instructions; read them together 
and apply them as follows:

\begin{itemize}[leftmargin=*]
    \item When the task feedback and the regularization feedback point to the 
    same part of the prompt, apply them as one combined edit.
    \item When they target different parts, apply both.
    \item When they conflict (e.g., the task feedback asks you to add a new 
    rule while the regularization feedback asks you to merge or shorten), 
    address the task feedback in the form most consistent with what the 
    regularization feedback specifies — by folding the task fix into the change 
    that the regularization feedback proposes, so the two are realized through 
    a single coordinated edit instead of two unrelated ones.
\end{itemize}

If a specific task feedback item genuinely cannot be addressed without violating 
a regularization feedback item, prioritize that task feedback item and apply it 
completely, even if this means setting aside that specific regularization item. 
All other regularization items in \texttt{<REG\_FEEDBACK>} still apply normally 
to the rest of the prompt.
\end{tcolorbox}

\subsection{Regularization-Guided Prompt Update — User Message Skeleton}
The user-message skeleton is sent at every rewriting step. The
\texttt{\{reg\_section\}} placeholder is filled with the \texttt{<REG\_FEEDBACK>} 
block when $g_{\text{reg}} \neq \varnothing$, and replaced by an empty string 
otherwise.
\begin{tcolorbox}[
    enhanced, breakable,
    colframe=black!70, colback=yellow!5,
    boxrule=1pt, arc=4mm,
    left=2mm, right=2mm, top=1mm, bottom=1mm,
    title={Regularization-Guided Prompt Update User Message Skeleton}
]
\small
\paragraph{Prefix template}
\begin{verbatim}
Here is the role of the variable you will improve: 
<ROLE>{variable_desc}</ROLE>.

The variable is the text within the following span: 
<VARIABLE> {variable_short} </VARIABLE>

{reg_section}Here is the context and task feedback we got 
for the variable:

<CONTEXT>{variable_grad}</CONTEXT>
\end{verbatim}

\paragraph{Reg-feedback section (filled into \texttt{\{reg\_section\}} when 
$g_{\text{reg}} \neq \varnothing$)}
\begin{verbatim}
The following is semantic edit regularization feedback. 
Follow its structural directives to control prompt length 
and specificity:

<REG_FEEDBACK>{reg_feedback}</REG_FEEDBACK>
\end{verbatim}
\end{tcolorbox}

\end{document}